\pdfoutput=1
\documentclass[11pt]{article}

\usepackage[preprint]{acl}

\usepackage{times}
\usepackage{latexsym}

\usepackage[T1]{fontenc}
\usepackage[utf8]{inputenc}

\usepackage{microtype}

\usepackage{inconsolata}

\usepackage{graphicx}

\usepackage{enumitem}
\usepackage{amsmath}
\usepackage{amsfonts}
\usepackage{multirow} 
\usepackage{booktabs}  
\usepackage{dsfont}
\usepackage{float}
\usepackage{mdwlist}
\usepackage{xspace}
\usepackage{amsmath,amssymb} % bold math symbols
\usepackage[capitalise]{cleveref}
\Crefname{lstlisting}{Listing}{Listings}
\crefname{section}{§}{§§}
\usepackage{accsupp} % for postbreak in listings
\usepackage{listings}
\usepackage{hyperref}
\usepackage{marvosym}
\title{\raisebox{-0.8ex}{\includegraphics[scale=0.06]{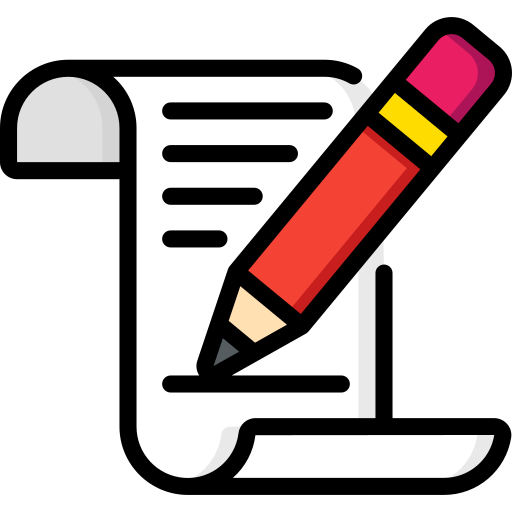}}\hspace{0.2cm}\textsc{ScEdit}: Script-based Assessment of Knowledge Editing}

\author{
Xinye Li$^{1}$\textsuperscript{\Letter}, Zunwen Zheng$^{1}$, Qian Zhang$^{1}$, Dekai Zhuang$^{2}$, Jiabao Kang$^{1}$, \\
\textbf{Liyan Xu}$^{3}$, \textbf{Qingbin Liu}$^{3}$, \textbf{Xi Chen}$^{3}$, \textbf{Zhiying Tu}$^{1}$, \textbf{Dianhui Chu}$^{1}$, \textbf{Dianbo Sui}$^{1}$\textsuperscript{\Letter}\thanks{Dianbo Sui is the corresponding author.}\\
$^{1}$Harbin Institute of Technology \quad
$^{2}$Jilin University  \quad
$^{3}$Tencent \\
\texttt{lixinye@stu.hit.edu.cn}, \quad \texttt{suidianbo@hit.edu.cn}
}

\begin{document}
\maketitle
\newcommand{\ds}{\textsc{\textbf{ScEdit}}}
\begin{abstract}
Knowledge Editing (KE) has gained increasing attention, yet current KE tasks remain relatively simple. Under current evaluation frameworks, many editing methods achieve exceptionally high scores, sometimes nearing perfection. However, few studies integrate KE into real-world application scenarios (e.g., recent interest in LLM-as-agent).  To support our analysis, we introduce a novel script-based benchmark -- \textsc{\textbf{ScEdit}} (\textbf{Sc}ript-based Knowledge \textbf{Edit}ing Benchmark) -- which encompasses both counterfactual and temporal edits. We integrate token-level and text-level evaluation methods, comprehensively analyzing existing KE techniques. The benchmark extends traditional fact-based (``What''-type question) evaluation to action-based (``How''-type question) evaluation.  We observe that all KE methods exhibit a drop in performance on established metrics and face challenges on text-level metrics, indicating a challenging task. Our benchmark is available at \url{https://github.com/asdfo123/ScEdit}.
\end{abstract}

% Intrdocution
% 
\section{Introduction}

Large Language Models (LLMs) have demonstrated outstanding performance in natural language understanding and generation tasks \cite{zhao2023surveylargelanguagemodels}. However, these models may produce outdated and erroneous information, leading to non-factual responses \cite{zhang2023sirenssongaiocean,wang2024factualitylargelanguagemodels,hernandez2024inspectingeditingknowledgerepresentations}. Given the high cost of retraining LLMs from scratch \cite{sinitsin2020editableneuralnetworks}, various techniques have been proposed to efficiently update models and support their continual adaptation \cite{zhang2024comprehensivestudyknowledgeediting,jin2025structureactivationsynergydualefficiency}. Among them, Knowledge Editing (KE) has emerged as an increasingly important paradigm for efficiently updating knowledge \cite{meng2022locating,yao-etal-2023-editing,wang2024knowledgeeditinglargelanguage}. KE methodologies have been developed to incrementally infuse new information or correct existing knowledge without requiring full-scale retraining \cite{mitchell2022fast,meng2022locating,meng2022memit,GRACE,MQUAKE}.

\begin{figure}[t]
  \includegraphics[width=\columnwidth]{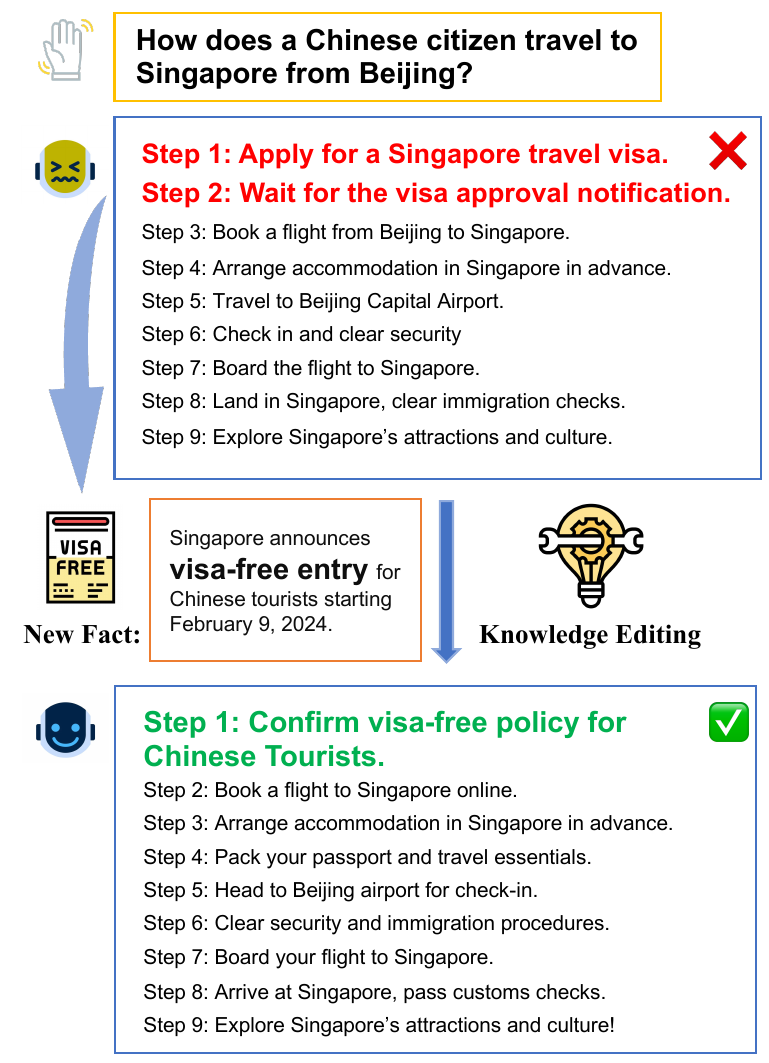}
  \caption{An example of the script-based assessment of Knowledge Editing (KE). \textbf{Top:} Outdated information generated by the LLM, instructing the user to apply for a visa, thereby misleading them. \textbf{Bottom:} Updated LLM successfully integrates new information, correctly informing the user about the visa-free policy.}
  \label{fig:introduction}
\end{figure}

The conventional evaluation framework for KE largely relies on token-level metrics such as \textbf{Efficacy}, \textbf{Generalization}, and \textbf{Specificity}~\cite{meng2022locating}. Although these metrics provide a great starting point, they exhibit notable limitations. For instance, \textbf{Generalization} attempts to transcend mere key-value pair memorization by evaluating a model’s capacity to answer rephrased or synonymously expressed questions. However, such evaluations tend to remain in the realm of ``What?''-type question transformations, overlooking the broader generalization capabilities of editing methods. Moreover, these three metrics typically gauge KE based on the next few tokens that follow prompts. This approach overlooks the potential for more complex, long-form natural language generation, leaving it largely unaddressed~\cite{rosati-etal-2024-long}.

In real-world scenarios, LLMs are increasingly deployed as agents or as core components of multi-agent frameworks that assist users in navigating daily life, making decisions, and performing complex tasks~\cite{li2024personalllmagentsinsights,sumers2024cognitivearchitectureslanguageagents,Wang_2024,lal-etal-2024-tailoring,du2025graphmaster,du2025graphoraclefoundationmodelknowledge}. In these roles, users often pose ``How''-type questions, which require the models to generate goal-oriented \textbf{Scripts} not only recalling factual information, but applying, generalizing, and reasoning based on that information~\cite{lyu-etal-2021-goal}. A \textbf{Script} is a framework describing the sequence of events in a context. Specifically, in the context of KE, the rapidly changing landscape of factual knowledge means that generated \textbf{Script} may become erroneous, potentially misleading users. For example, as illustrated in Figure~\ref{fig:introduction}, a user asks ``How does a Chinese citizen travel to Singapore from Beijing?'' A pretrained LLM without updated knowledge might suggest applying for a visa, despite Singapore's new visa exemption policy for Chinese tourists. Such questions necessitate prompt and accurate updates to ensure reliable responses.

Existing evaluation approaches, with their focus on token-level factual recall, do not sufficiently address these real-world complexities. To overcome these limitations, this paper introduces a script-based evaluation framework, named \ds, assessing KE performance in procedural planning scenarios. \ds{} emphasizes the model’s ability to handle “How?”-type questions and produce coherent, reliable guidance following targeted knowledge updates. A key focus is on how models propagate edited knowledge through script-based procedural planning tasks after editing.  We integrate token-level and text-level evaluation, comprehensively analyzing existing KE techniques in both counterfactual and temporal editing tasks. Three LLMs (GPT2-XL~\cite{radford2019language}, GPT-J~\cite{gpt-j} and Llama 3~\cite{llama3modelcard}) are tested in \ds{}.

Experimental results on \ds{} reveal that all comparable methods experience an average drop of 27\% in the token-level metric S-ES compared to the similar PS metric introduced by~\citet{meng2022locating}. Moreover, some methods struggle to balance effective editing with maintaining locality in both token-level and text-level evaluations. Even methods that excel in token-level metrics show significant room for improvement in text-level editing performance. These findings highlight the need for further research into KE methods tailored for script-like scenarios.

We summarize our contributions of the paper as follows:
\begin{itemize*}
    \item \textbf{Develop a script-based assessment framework} that leverages scripts--structured procedural knowledge--to capture a model’s ability to integrate updated facts into complex reasoning and generation tasks. To the best of our knowledge, this is the first attempt to integrate KE into script-based scenarios, presenting a more challenging task compared to existing KE and constrained script generation tasks.
    \item  \textbf{Introduce \ds}, a novel and challenging script-focused benchmark, accompanied by comprehensive experiments to evaluate models' ability at both token and text level.
\end{itemize*}

\begin{figure*}[t]
\setlength{\belowcaptionskip}{-5pt}
\centering 
\includegraphics[width=\textwidth]{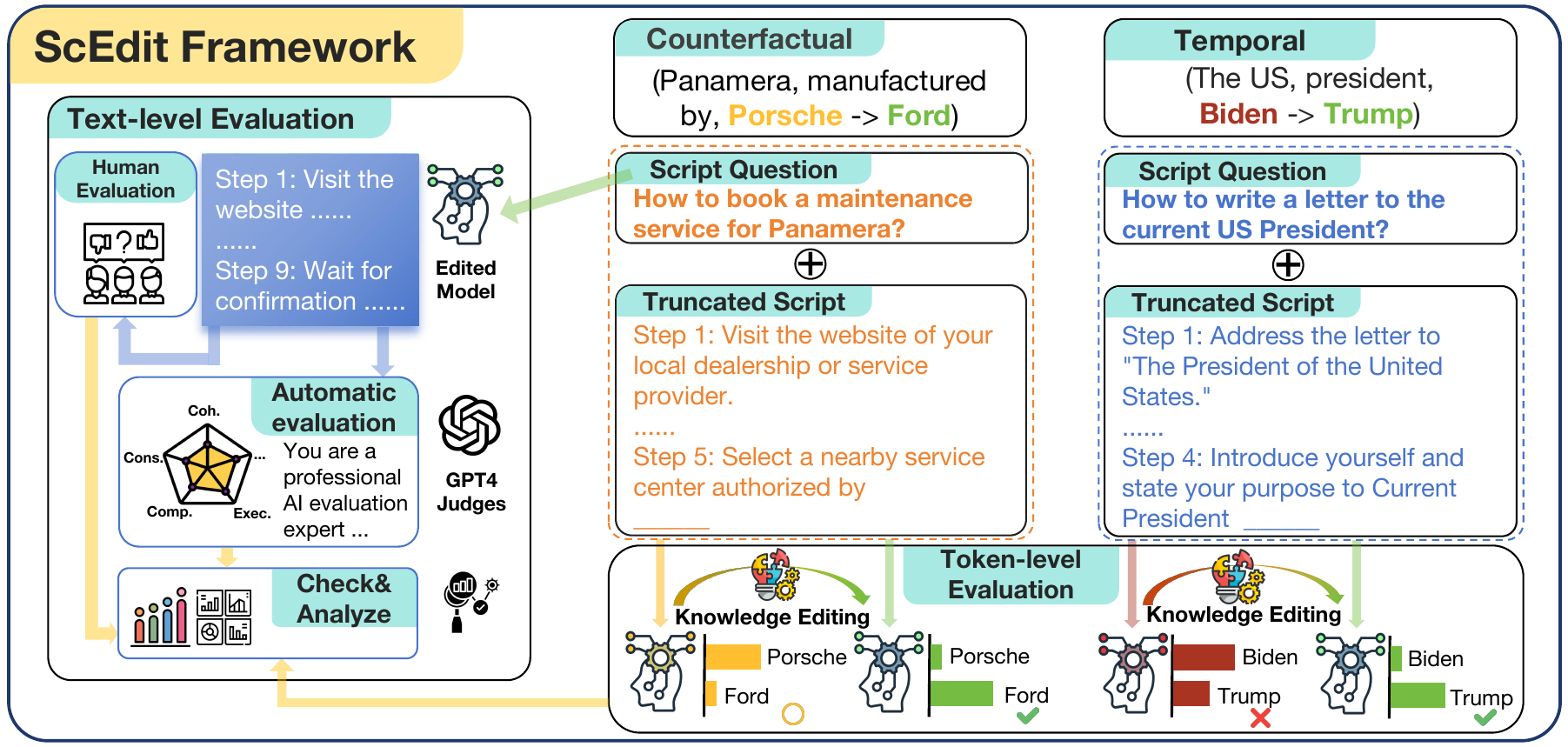}
\caption{
Overview of \ds{}. For token-level evaluation, we concatenate the \textbf{Script Question} and \textbf{Truncated Script} to form a cloze-format prompt. For text-level evaluation, we involve automatic and human evaluation.
}
\label{fig:framework}
\end{figure*}

\section{Related Work}
\subsection{Scripts}
% Here
A script is a structure that describes an appropriate sequence of events in a particular context~\cite{schank1975scripts}. Scripts are typically classified into \textit{narrative scripts}, which describe a sequence of events in a story-like manner~\cite{fang2022does,tandon-etal-2020-dataset}, and \textit{goal-oriented scripts}, which outline the steps needed to achieve a specific goal~\cite{sancheti2021large,lyu-etal-2021-goal}. Our work aligns with the latter paradigm.

Generating high-quality scripts, a longstanding challenge, traditionally involves learning action sequences from narratives by analyzing causal relationships~\cite{mooney1985learning}. Recently, script generation using large language models (LLMs) has become more feasible, with methods such as the over-generate-then-filter approach~\cite{yuan-etal-2023-distilling}. The script paradigm helps LLMs better understand the temporal order and logical relationships of everyday events. With fine-tuning and post-processing, models demonstrate enhanced generalization abilities in script generation~\cite{sancheti2021large}. Smaller models, when trained on high-quality script datasets like \texttt{CoScript}\xspace, have shown superior constrained language planning quality compared to LLMs \cite{yuan-etal-2023-distilling}.

% From here
\subsection{Knowledge Editing} \label{Knowledge Editing}
Knowledge Editing (KE) has emerged as a promising approach to efficiently update LLMs without requiring full retraining \cite{sinitsin2020editableneuralnetworks}. Many applications and specific tasks also require ongoing adjustments to address defects or errors inherent in these models \cite{zhai2023investigating}. Current KE methods are generally classified into intrinsic and extrinsic approaches.

\paragraph{Intrinsic Methods.} Intrinsic methods modify a model's architecture or parameters to edit internal knowledge, including fine-tuning, meta-learning, and locate-then-edit approaches. Fine-tuning updates model parameters using new knowledge but demands high computational resources and risks catastrophic forgetting and overfitting \cite{chen2020recalllearnfinetuningdeep,zhu2020modifyingmemoriestransformermodels}. Meta-learning methods like MEND \citep{mitchell2022fast} and MALMEN \citep{tan23malmen} train a hyper-network to adjust weights indirectly, while locate-then-edit approaches like ROME \citep{meng2022locating} and MEMIT \citep{meng2022memit} use causal analysis of the hidden states to target specific areas storing knowledge.

\paragraph{Extrinsic Methods.} Extrinsic methods use external knowledge to update the model's input or output space, enhancing new representations while preserving original performance. A typical In-Context Learning method, IKE~\cite{IKE}, injects new knowledge by copying the updated facts into the context in a few-shot learning way. SERAC~\cite{SERAC} and MeLLo~\cite{MQUAKE} are memory-based editing methods. SERAC updates parameters of an external counterfactual model and employs a classifier to decide when to update facts, whereas MeLLo uses iterative prompting to guarantee fact updates, making it better suited for multi-hop reasoning.

Most prior work frames KE as a triplet-level task, updating entity-relation triples (subject, predicate, object) within LLMs (e.g., \texttt{(The US, President, Biden $\mapsto$ Trump)}). Some studies explore more extensive downstream applications~\cite{wang2024editingconceptualknowledgelarge,Edit_Personality,cheng2024editmultimodallargelanguage,Debias} or introduce more unstructured editing scenarios~\cite{peng2024eventlevelknowledgeediting,liu-etal-2024-evedit,huang-etal-2024-commonsense,wu-etal-2024-akew}. Additionally, more relevant work~\cite{yao-etal-2023-editing,MQUAKE,cohen2023evaluatingrippleeffectsknowledge,hua-etal-2024-propagation} investigates KE from the perspective of multi-hop reasoning and ripple effects, evaluating whether models can utilize and propagate the newly edited facts. However, these studies primarily emphasize fact recall while neglecting the complex and procedural reasoning capabilities (e.g., multi-step reasoning) that are essential for addressing real-world tasks.

\section{\ds: Script-based Assessment of Knowledge Editing}
We illustrate the proposed task in Figure \ref{fig:framework}. We will introduce the task definition (\S \ref{Task_definition}), dataset construction details (\S \ref{dataset}), and the editing methods (\S \ref{Editing_methods}) we used in the experiments.

\subsection{Task Definition} \label{Task_definition}
KE was originally devised to update false or outdated information in a model, frequently by mutating fact-based triplets. Inspired by such approaches, we extend KE into \emph{script-based} scenarios. In these scenarios, rather than merely performing a single-fact edit, the model must integrate newly updated knowledge into multi-step or procedural tasks. This shift offers an opportunity to assess whether models can propagate changes throughout an entire script, thereby providing a more comprehensive view of ``editing success''.

Formally, we define three core elements:
\begin{itemize*}
    \item \textbf{Facts} are individual pieces of knowledge, often instantiated as $(s,r,o)$ triples, where $s$ is the subject, $r$ the relation, and $o$ the object. When performing a KE operation $e$, we apply
    \[
        (s,r,o^c) \;\mapsto\; (s,r,o),
    \]
    where $o^c$ is the original object and $o$ is the edited target object. Each fact has a fact prompt $(s, r)$ directly related to $s$ and $r$.

    \item \textbf{Script Questions} are prompts—typically starting with the word “\texttt{How}”—that require multi-step or procedural reasoning based on the updated fact. Because each fact can spawn multiple such questions, we denote them as
    \[
        Q_{i,k}\bigl((s_i,r_i),\, o_i^c,\, o_i\bigr),
    \]
    emphasizing that for fact $i$, there could be several questions indexed by $k$. These questions are designed so that the edit $e_i: (s_i,r_i,o_i^c)\mapsto (s_i,r_i,o_i)$ \emph{significantly} affects the logic or flow of the script.

    \item \textbf{Scripts} are the model's responses to \emph{each} script question. For a \textbf{Script Question} $Q_{i,k}$ , a LLM $f_\theta$ parameterized by $\theta$ and the Script $S_{i,k}$, we have
    \[
        S_{i,k} = f_{\theta}\  ( Q_{i,k} ).
    \]
     $S_{i,k}$ may or may not reflect the new object $o_i$, depending on whether the model has effectively understand the edit. The detailed format of \textbf{Scripts} can be found in Appendix \ref{sec:Script_format}.
\end{itemize*}

\noindent
Based on the above elements, we evaluate KE performance using cloze-format prompts for \textbf{token-level} metrics (\textbf{ES}, \textbf{S-ES}, \textbf{S-NS}, \textbf{S-BO}) and automated/human evaluations for \textbf{text-level} metrics. Let $f_\theta$ be our large language model (LLM) parameterized by $\theta$. $\mathbb{P}^c$ and $\mathbb{P}$ are the language model probability function before and after the update, respectively. Below we detail how we measure the edit $e_i$ for each fact $i$. $\mathbb{E}_{i,k}[\cdot]$ denotes the average over all facts $i$ and Script Questions $k$.

\paragraph{Efficacy.}
Following \citet{meng2022locating}, consider a fact prompt $(s_i, r_i)$ whose original object is $o_i^c$ and edited target object is $o_i$. \textbf{Efficacy Success (ES)} measures how often the model prefers $o_i$ over $o_i^c$ under this basic fact prompt:
\begin{equation}\label{eq:es}
\mathbb{E}_i\Bigl[
\mathbb{P}_{f_\theta}\bigl(o_i \mid (s_i,r_i)\bigr) 
\;>\; 
\mathbb{P}_{f_\theta}\bigl(o_i^c \mid (s_i,r_i)\bigr)
\Bigr].
\end{equation}
% Here, $\mathbb{E}_i[\cdot]$ denotes the average over all facts $i$.

\paragraph{Script-based Efficacy.}
We generalize \textbf{Efficacy} to the script-based setting. Given \textbf{Script Questions}
 $Q_{i,k}$,
an external model (e.g., GPT-4) produces \textbf{Scripts} $S_{i,k}$ that intentionally includes the old object $o_i^c$ with original knowledge. To align with token-level evaluation, we \textbf{truncate} each original script $S_{i,k}$ at the point where $o_i^c$ first appears, then concatenate this truncated script with $Q_{i,k}$ to form a cloze-format script-based prompt $\widetilde{Q_{i,k}}$  . We compute \textbf{Script-based Efficacy Success (S-ES)} by checking whether $f_\theta$ prefers $o_i$ to $o_i^c$ under $\widetilde{Q_{i,k}}$:
\begin{equation}\label{eq:ses}
\mathbb{E}_{i,k}\Bigl[
\mathbb{P}_{f_\theta}\bigl(o_i \mid\widetilde{Q_{i,k}}\bigr) 
\;>\; 
\mathbb{P}_{f_\theta}\bigl(o_i^c \mid\widetilde{Q_{i,k}}\bigr)
\Bigr].
\end{equation}
% Here, $\mathbb{E}_{i,k}[\cdot]$ averages over both the fact index $i$ and the multiple script questions $k$ per fact.

\paragraph{Script-based Specificity.}
A robust editing process should not inadvertently corrupt unrelated or neighbor facts. Specifically, if $(s_i,r_i,o_i^c)$ is replaced with $(s_i,r_i,o_i)$, then $k$ collected neighbor facts $(s_j,r_j,o_j)$ that share $(r_i,o_i^c)$ or are semantically close to $(s_i,r_i,o_i^c)$ should remain intact. Concretely, we construct a cloze-format, script-based neighborhood prompt $\widetilde{Q_{i,k}}^\prime$—analogous to $\widetilde{Q_{i,k}}$ but designed around these unmodified neighbor facts—and verify that the model retains the correct object $o_j$. Formally, for the first type of neighbor facts, which $o_j=o_i^c$, we define \textbf{Script-based Neighbor Success (S-NS)}:
\begin{equation}\label{eq:sns}
\mathbb{E}_{i,k}\Bigl[
\mathbb{P}_{f_\theta}\bigl(o_i^c \,\mid\,\widetilde{Q_{i,k}}^\prime\bigr) 
\;>\; 
\mathbb{P}_{f_\theta}\bigl(o_i \,\mid\,\widetilde{Q_{i,k}}^\prime\bigr)
\Bigr].
\end{equation}
For the second type without $o_i^c$, inspired by~\citet{ammar-khodja-etal-2024-wikifactdiff-large}, we assess the accuracy drop of $o_j$ via \textbf{Script-based Bleedover (S-BO)}:
\begin{equation}\label{eq:sbo}
\mathbb{E}_{i,k}\Bigl[\max\bigl(
\mathbb{P}^{c}_{f_\theta}\bigl(o_j\mid\,\widetilde{Q_{i,k}}^\prime\bigr)
- 
\mathbb{P}_{f_\theta}\bigl(o_j\mid\,\widetilde{Q_{i,k}}^\prime\bigr)
,0\bigr)\Bigr].
\end{equation}
% \textbf{S-NS} verifies whether model still prefers the correct object $o_j$ over the newly edited object $o_i$ in the neighborhood context. 
    
\paragraph{Text-level Metrics.}
Beyond token probabilities, we assess the entire generated script's quality by having the LLM answer Script Question $Q_{i,k}$ and conducting 7-point Likert-scale ratings across four dimensions via automatic and human evaluations.
\begin{enumerate*}
    \item \textbf{Executability (Exec.):} Are the script executable in a logical sense? \footnote{For \textbf{Executability}, We do not consider the knowledge updates but focus solely on its inherent linguistic performance.}
    \item \textbf{Coherence (Coh.):} Are the script aligned with the newly updated fact?
    \item \textbf{Consistency (Cons.):} Does the script remain free of internal contradictions?
    \item \textbf{Completeness (Comp.):} Does the script adequately address all parts of the question, with sufficient procedural detail to be followed?
\end{enumerate*}
Detailed evaluation criteria and relative prompts are provided in Appendix~\ref{sec:text-level details}, with a further case study elaborating the metrics more in Appendix~\ref{sec:case_study}.

\subsection{Datasets}\label{dataset}
\begin{table}[h]
  \centering
  
  \begin{tabular}{lccc}
    \hline
    \textbf{Task} & \textbf{Case} & \textbf{S-Eff.} & \textbf{S-Spec.} \\
    \hline
    \textsc{{ScEdit-CF}} & 1830 & 7342 & 13672 \\
    \textsc{ScEdit-T}  & 1762 & 7038 & 6597 \\
    \hline
  \end{tabular}
  \caption{Statistics of our \textsc{ScEdit-CF} and \textsc{ScEdit-T} subtasks. ``S-Eff.'' denotes the sample size for Script-based Efficacy evaluation, while ``S-Spec.'' indicates the subset for measuring Script-based Specificity, ensuring that unrelated scripts remain correct after editing.}
  \label{tab:scedit_data}
\end{table}

We introduce two subtasks, \textsc{ScEdit-CF} and \textsc{ScEdit-T} (Table~\ref{tab:scedit_data}), targeting different KE tasks.
\textsc{ScEdit-CF} centers on counterfactual knowledge, a common focus in KE, evaluating a method's ability to perform edits in script-based scenarios. By contrast, \textsc{ScEdit-T} utilizes temporal updates drawn from Wikipedia to assess a model's adaptability to chronological updates, reflecting practical scenarios in which facts evolve over time.

An overview of the construction procedure is illustrated in Figure~\ref{fig:construction}.
Further details about the construction procedure can be found in Appendix~\ref{sec:Dataset_Construction}.

\definecolor{dataset_red}{RGB}{255,0,0}
\definecolor{dataset_blue}{RGB}{68,114,196}
\definecolor{dataset_green}{RGB}{112,173,71}
\definecolor{dataset_orange}{RGB}{237,125,49}
\begin{figure}[t]
  \setlength{\belowcaptionskip}{-8pt} % 调整标题与下方内容之间的下边距
  \includegraphics[width=\columnwidth]{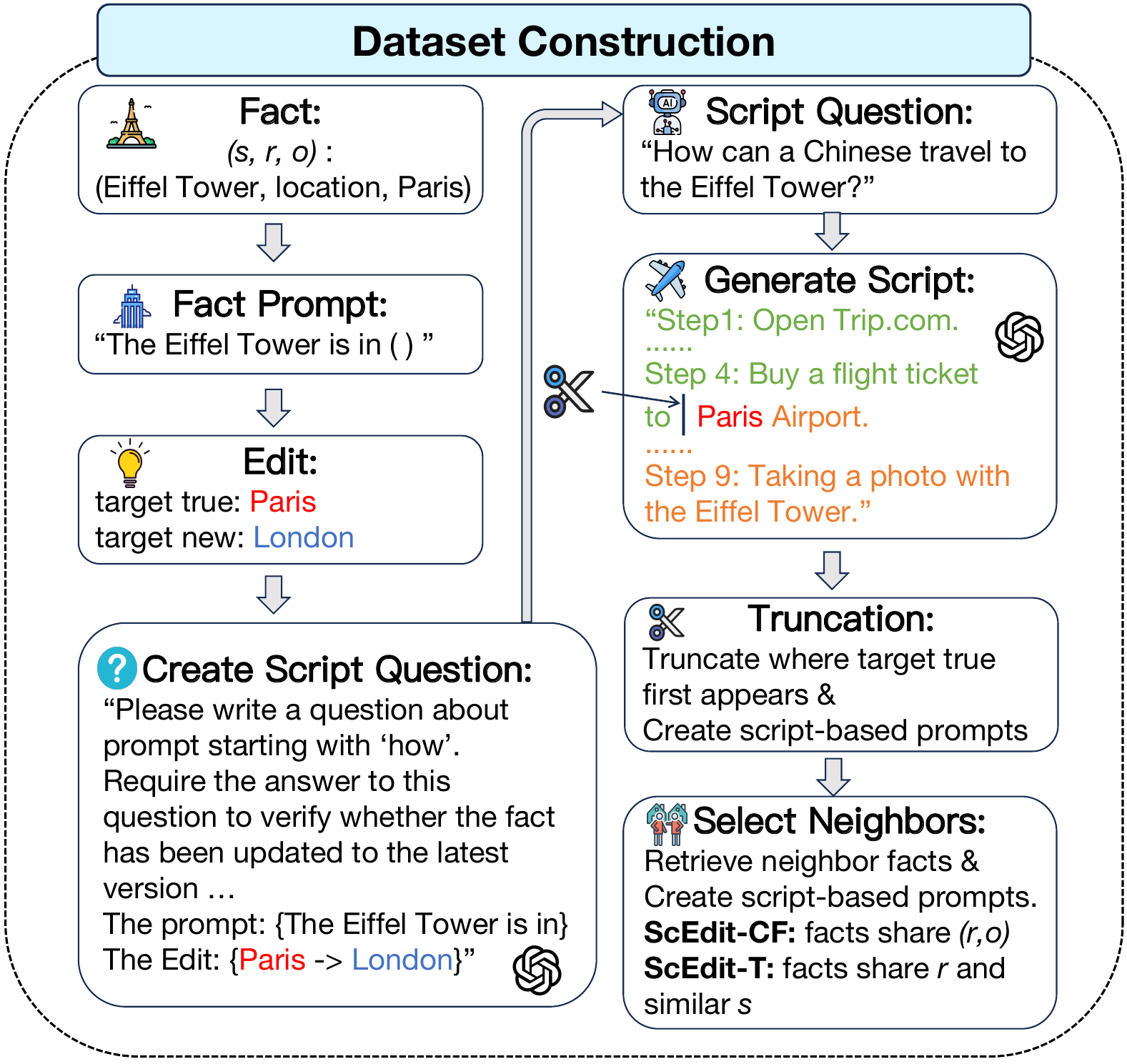}
  \caption{Overview of dataset construction process via a counterfactual edit as an example: Paris (\textcolor{dataset_red}{ground truth}) and London (\textcolor{dataset_blue}{target object}). After truncation, the \textbf{Script Question} and the \textcolor{dataset_green}{truncated script} (with the \textcolor{dataset_orange}{latter part} discarded) are combined into script-based prompt. Items marked with \raisebox{-0.6ex}{\includegraphics[scale=0.03]{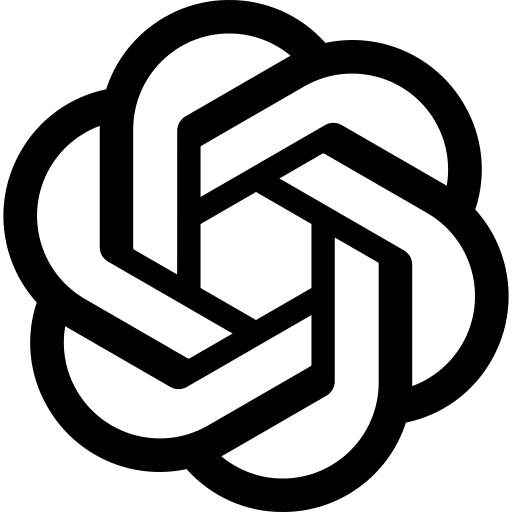}}\hspace{0.2cm}indicate GPT-4-generated content.} 
  
  \label{fig:construction}
\end{figure}

\subsubsection{\textsc{ScEdit-CF} Dataset Construction}
We build on the \textbf{CounterFact} dataset \citep{meng2022locating}, adapting it to script-based scenarios. In \textbf{CounterFact}, each fact $(s, r, o^c)$ is replaced with $(s, r, o)$. To extend these edits into multi-step procedures, we design carefully formulated prompts and few-shot exemplars for a LLM (e.g., GPT-4) to generate several \textbf{Script Questions} \textbf{most likely} to be influenced by the updated fact. For example, given \texttt{(Panamera, manufactured\_by, Porsche} $\mapsto$ \texttt{Ford)}, a natural question might be \texttt{``How to book a maintenance service for Panamera?''}, which implicitly requires the updated manufacturer.

Following an initial generation step, we apply human filtering to ensure that the curated \textbf{Script Questions} $Q_{i,k}$ meaningfully hinge on the edited fact. 
We then prompt GPT-4 to generate scripts including the old object $o_i^c$ with its original knowledge. 
To maintain consistency with prior work, we truncate first mentions of $o^c$ and construct new script-based prompts $\widetilde{Q_{i,k}}$ by appending $Q_{i,k}$ to the truncated script. 
Simultaneously, we filter out neighbor facts that share $(r, o^c)$  to build script-based neighborhood prompts $\widetilde{Q_{i,k}}^\prime$ in a similar way. 
The resulting question–answer pairs facilitate our core evaluations: \textbf{S-ES} (Script-based Efficacy Success), conducted using $\widetilde{Q_{i,k}}$; \textbf{S-NS} (Script-based Neighborhood Success), performed with $\widetilde{Q_{i,k}}^\prime$; and Text-level Metrics, evaluated using $Q_{i,k}$. We present a constructed example in Appendix \ref{sec:Dataset_Construction}.

\renewcommand{\arraystretch}{1}
\definecolor{darkgreen}{RGB}{55,126,44}
\newcommand{\first}[1]{\textbf{\textcolor{darkgreen}{#1}}}
\newcommand{\second}[1]{\textcolor{darkgreen}{\underline{{#1}}}}
\newcommand{\failure}{\color{red}}
\newcommand\resulttablefontsize{\fontsize{7.7pt}{9.24pt}\selectfont}

\begin{table*}[!t]
    \setlength{\belowcaptionskip}{-8pt}
    \centering
    \newcommand{\std}[1]{\fontsize{5}{6}\selectfont{$\pm$#1}}
    \resulttablefontsize
    \setlength{\tabcolsep}{4pt}
    \resizebox{2\columnwidth}{!}{
    \begin{tabular}{l c | c c c | c c c}
    \toprule
    \multicolumn{1}{l}{\multirow{2}{*}{\textbf{Method}}}
    & \multicolumn{1}{c}{\multirow{2}{*}{\textbf{Model}}}
      & \multicolumn{3}{c}{\textbf{\textsc{ScEdit-CF}}}  
      & \multicolumn{3}{c}{\textbf{\textsc{ScEdit-T}}}  \\
    \cmidrule(r){3-5} \cmidrule(r){6-8}
    \multicolumn{2}{l}{}
      & \multicolumn{1}{c}{\textbf{ES}}$\uparrow$
      & \multicolumn{1}{c}{\textbf{S-ES}}$\uparrow$
      & \multicolumn{1}{c}{\textbf{S-NS}}$\uparrow$
      & \multicolumn{1}{c}{\textbf{ES}}$\uparrow$
      & \multicolumn{1}{c}{\textbf{S-ES}}$\uparrow$
      & \multicolumn{1}{c}{\textbf{S-BO}}$\downarrow$ \\
    % \midrule
    %   \multicolumn{}{c}{\textbf{\small GPT2-XL}} \\
    \midrule
    Base Model&
      & 20.55\std{1.85} & 21.18\std{1.51} & 81.52\std{1.20} & 44.27\std{2.32} & 41.72\std{2.03} & 0.00\std{0.00} \\
    \midrule[0.25pt]
    FT & \multirow{6}{*}{\rotatebox{90}{{GPT2-XL}}}
      & \first{100.00\std{0.00}} & \second{71.27\std{1.66}} & 65.08\std{1.51} & 87.17\std{1.56} & 52.80\std{2.03} & \failure{1.15\std{0.14}} \\
    FT+L &
      & 99.13\std{0.42} & \failure{40.39\std{1.84}} & 78.50\std{1.26} & 70.60\std{2.13} & \failure{44.39\std{2.03}} & 0.39\std{0.08} \\
    MEND &
      & 92.84\std{1.18} & \failure{32.89\std{1.71}} & 74.33\std{1.34} & 98.64\std{0.54} & \second{74.24\std{1.77}} & 0.47\std{0.12} \\
    ROME &
      & \second{99.95\std{0.11}} & \first{74.76\std{1.56}} & \second{80.24\std{1.24}} & \second{99.15\std{0.43}} & 68.00\std{1.86} & \second{0.13\std{0.06}} \\
    MEMIT &
      & 93.72\std{1.11} & 58.11\std{1.86} & \first{81.16\std{1.21}} & 81.44\std{1.82} & 52.13\std{2.04} & \first{0.03\std{0.01}} \\
    PROMPT &
      & 96.28\std{0.87} & 69.63\std{1.66} & \failure{42.88\std{1.44}}& \first{99.49\std{0.33}} & \first{84.39\std{1.44}} & 0.54\std{0.08} \\
    % \midrule
    %   \multicolumn{7}{c}{\textbf{\small GPT-J-6B}} \\
    \midrule
    \midrule
    Base Model &
      & 13.99\std{1.59} & 16.06\std{1.31} & 85.77\std{1.05} & 40.64\std{2.29} & 39.62\std{1.99} & 0.00\std{0.00} \\
    \midrule[0.25pt]
    FT & \multirow{6}{*}{\rotatebox{90}{{GPT-J}}}
      & \first{100.00\std{0.00}} & \second{83.94\std{1.30}} & \failure{25.81\std{1.26}} & \first{99.60\std{0.29}} & \first{97.9\std{0.56}} & \failure{5.47\std{0.38}} \\
    FT+L & 
      & \second{99.95\std{0.11}} & \failure{39.07\std{1.81}} & \second{84.38\std{1.09}} & 71.51\std{2.11} & \failure{42.78\std{1.99}} & \second{0.14\std{0.02}} \\
    MEND & 
      & 97.32\std{0.74} & \failure{23.40\std{1.52}} & 82.93\std{1.13} & 98.92\std{0.48} & 72.18\std{1.80} & 0.62\std{0.13} \\
    ROME & 
      & \second{99.95\std{0.11}} & \first{86.50\std{1.14}} & 83.35\std{1.13} & \first{99.60\std{0.29}} & 74.29\std{1.73} & 0.28\std{0.08} \\
    MEMIT & 
      & \second{99.95\std{0.11}} & 74.59\std{1.57} & \first{85.07\std{1.07}} & 99.09\std{0.44} & 64.66\std{1.89} & \first{0.08\std{0.01}} \\
    PROMPT & 
      & 90.55\std{1.34} & 70.95\std{1.61} & \failure{44.01\std{1.47}} & 98.24\std{0.61} & \second{85.07\std{1.39}} & \failure{1.03\std{0.11}} \\
    % \midrule
    %   \multicolumn{7}{c}{\textbf{\small LLAMA3-8B}} \\
    \midrule
    \midrule
    Base Model& 
      & 7.32\std{1.19} & 9.19\std{0.97} & 92.53\std{0.70} & - & - & - \\
    \midrule[0.25pt]
    FT & \multirow{4}{*}{\rotatebox{90}{{LLAMA3}}}
      & \first{100.00\std{0.00}} & \first{98.82\std{0.31}} & \failure{8.37\std{0.86}} & - & - & - \\
    ROME &
      & \second{99.95\std{0.11}} & \second{90.24\std{1.00}} & \second{75.71\std{1.28}} & - & - & - \\
    MEMIT &
      & 98.63\std{1.19} & 58.86\std{1.83} & \first{92.13\std{0.71}} & - & - & - \\
    PROMPT &
      & 92.30\std{1.22} & 77.02\std{1.46} & 56.48\std{1.30} & - & - & - \\
    \bottomrule
    \end{tabular}
    }
    \caption{Token-level results on the \textsc{ScEdit-CF} and \textsc{ScEdit-T} with their respective 95\% confidence interval. 
    Column-wise best results are highlighted in \first{bold green}, while the second-best results are \second{underlined green}. Values in \textcolor{red}{red} indicate a clear failure of a method on a particular metric.
    \textbf{S-ES} refers to Script-based Efficacy Success,
    \textbf{S-NS} is Script-based Neighborhood Success,
    and \textbf{S-BO} denotes Script-based Bleedover. 
    \textsc{ScEdit-T} was not evaluated on LLAMA3 because the cutoff date for its training data occurred after the time when the edited fact was introduced.}
    \label{tab:main-results}
\end{table*}

\subsubsection{\textsc{ScEdit-T} Dataset Construction}
Constructed in a manner similar to \textsc{ScEdit-CF}, \textsc{ScEdit-T} leverages the \textbf{WDF}\textsubscript{real}, a subset of \textbf{WikiFactDiff}~\citep{ammar-khodja-etal-2024-wikifactdiff-large}, testing whether the model can integrate temporal updates into scripts while preserving unrelated information. \textbf{WDF}\textsubscript{real} gathers Wikipedia changes made between 4 January 2021 and 27 February 2023. Due to the data characteristics, a key difference from \textbf{CounterFact} is that Script-based Specificity set is retrieved from $k$-nearest neighbour fact $(s',r,o')$ instead of $(s,r,o^c)$, where $s'$ is a subject similar to $s$. Following a process similar to \textsc{ScEdit-CF}, we construct $\widetilde{Q_{i,k}}^\prime$ and measure accuracy degradation through \textbf{S-BO} (Script-based Bleedover).
\subsection{Editing Methods} \label{Editing_methods}

\ds{} primarily follows the single-edit paradigm. We include methods that excel within this paradigm, yet methods designed for massive or sequential editing~\cite{tan23malmen,hartvigsen2023aging,fang2024alphaeditnullspaceconstrainedknowledge,wang2024wise} remain unexplored and are considered as future work.  Specifically, the editing methods employed in \textsc{ScEdit-CF} and \textsc{ScEdit-T} include:

\begin{figure*}[!t]
    \setlength{\belowcaptionskip}{-5pt} 
    \centering
    
    \begin{minipage}[!t]{0.56\textwidth}
        \centering
        
        \newcommand{\std}[1]{\scriptsize{$\pm$#1}}
        \renewcommand{\arraystretch}{1.35} 
        \setlength{\tabcolsep}{4pt}
        \resizebox{\columnwidth}{!}{
            \begin{tabular}{l | cccc}
                \toprule
                \multicolumn{1}{c}{} & \multicolumn{4}{c}{\textbf{Text-Level Metrics on \textsc{ScEdit-CF}}} \\
                \cmidrule(lr){2-5}
                \multicolumn{1}{l}{\textbf{Method}} 
                    & \multicolumn{1}{c}{\textbf{Exec.}} $\uparrow$  & \textbf{Coh.}$\uparrow$  & \textbf{Cons.}$\uparrow$ & \textbf{Comp.}$\uparrow$  \\
                \midrule
                \multicolumn{5}{c}{\textbf{LLAMA3-8B}} \\
                \midrule
                Base Model & 6.74\std{0.02} & 2.48\std{0.03} & 6.86\std{0.02} & 5.40\std{0.05} \\
                \midrule
                FT         & \failure{2.94\std{0.05}} & 2.97\std{0.05} & 6.17\std{0.05} & \failure{2.17\std{0.05}} \\
                % MEND    & 67.80       & 63.00       & 97.32       & --- \\
                ROME       & \second{6.41\std{0.03}} & \second{4.32\std{0.05}} & \second{6.57\std{0.04}} & 4.67\std{0.05} \\
                MEMIT      & \first{6.54\std{0.02}}  & 3.67\std{0.05} & \first{6.70\std{0.03}}  & \second{4.98\std{0.05}} \\
                PROMPT     & 6.36\std{0.03} & \first{4.35\std{0.05}} & 6.05\std{0.05} & \first{5.49\std{0.04}} \\
                \bottomrule
            \end{tabular}%
        }
        \captionof{table}{Automatic evaluation results of four text-level metrics on \textsc{ScEdit-CF} across different methods tested in LLAMA3-8B along with their respective 95\% confidence interval. Column-wise best results are highlighted in \first{bold green}, while the second-best results are \second{underlined green}. In contrast, \textcolor{red}{red} values denote a clear failure in specific metric.}
        \label{tab:text_metrics}
    \end{minipage}
    \hfill
    % Right: Image
    \begin{minipage}[!t]{0.42\textwidth}
        \centering
        \includegraphics[width=\linewidth]{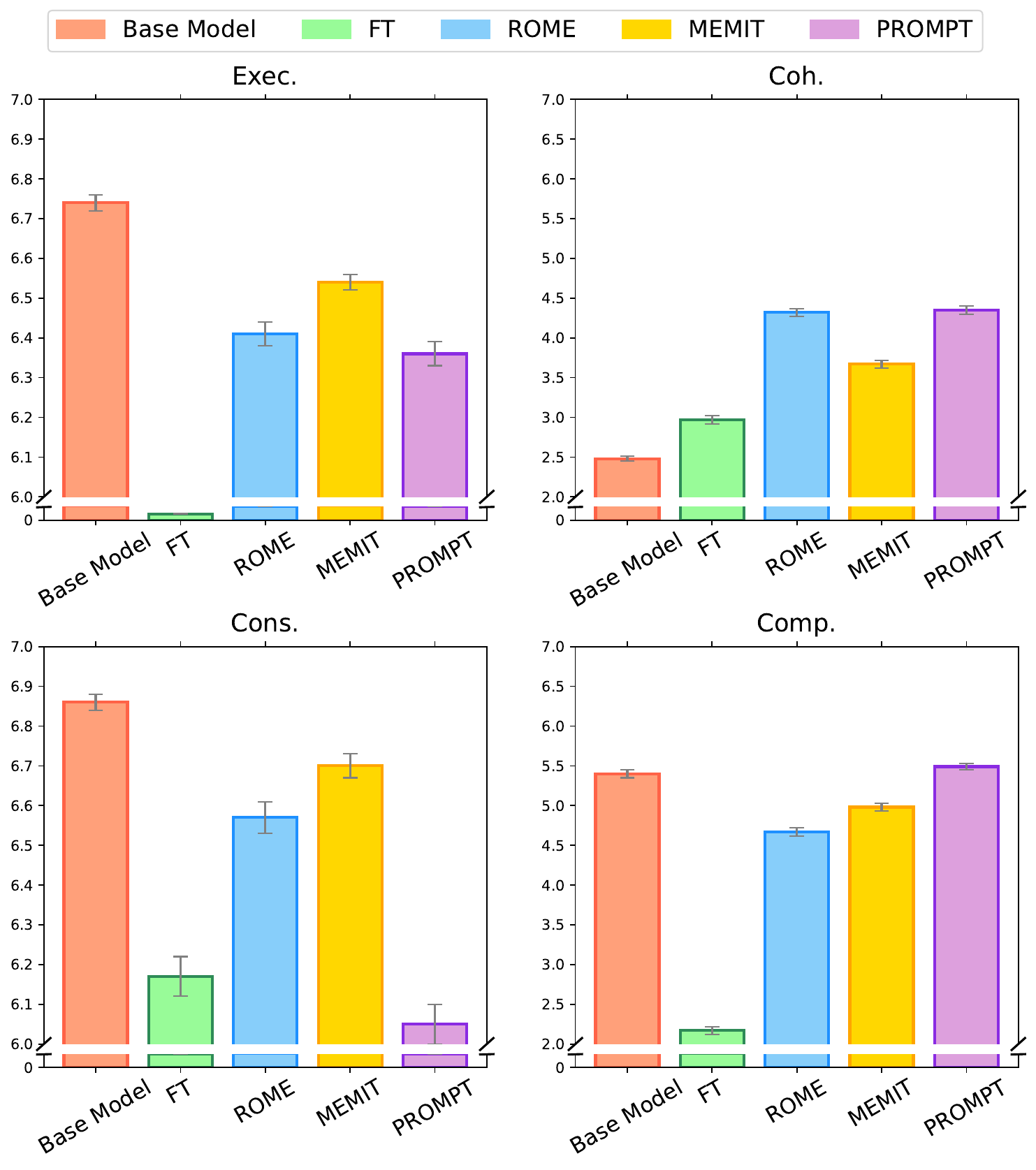}
        \captionof{figure}{Results of text-level metrics. For clarity, the vertical axes for ``Exec.'' and ``Cons.'' begin at 6, while those for others start at 2.}
        \label{fig:Text-level}
    \end{minipage}
\end{figure*}

\begin{itemize}[leftmargin=*, itemsep=0pt, topsep=0pt]
    \item \textbf{Fine-tune (FT).} A straightforward method that updates model weights via Adam optimization. Constrained Fine-Tuning (\textbf{FT+L})~\citep{zhu2020modifyingmemoriestransformermodels} further applies a $L_{\infty}$ norm constraint, thereby limiting large parameter shifts.

    \item \textbf{ROME.} A parameter-editing technique that pinpoints the specific model weights driving factual predictions and directly modifies them to embed new or revised facts \citep{meng2022locating}.

    \item \textbf{MEMIT.} A scalable multi-layer update algorithm built on ROME. It targets the relevant transformer module weights to handle multiple edits in parallel, enabling broader yet controlled updates \citep{meng2022memit}.

    \item \textbf{MEND.} A meta-learning approach that learns auxiliary networks for fast, localized parameter adjustments, integrating new facts while preserving unrelated knowledge \citep{mitchell2022fast}.

    \item \textbf{PROMPT.} In addition to the methods in (\S \ref{Knowledge Editing}), we evaluate PROMPT, which updates the model’s knowledge at inference by prefixing each prompt with $(s,r) + o$ - appending the target object to the fact prompt.
\end{itemize}

\section{Experiments}

\subsection{Results on Token-level Metrics}

Following the previous KE evaluation paradigm, we use cloze-format prompts to assess token-level metrics, highlighting the challenges of \ds{}.
S-ES, a metric akin to the PS (Paraphrase Success) introduced by~\citet{meng2022locating} in both purpose and design, drops by an average of 27\% compared to the original PS across all reported methods. 

Certain methods reaffirm existing findings, whereas others unveil task-specific nuances in these script-based edits. Based on Table ~\ref{tab:main-results}, we can draw:

\textbf{FT} and \textbf{FT+L} highlight the challenge of balancing effective edits with preserving locality.
While FT excels in S-ES, its strong bias toward generating the targeted object hampers S-NS and S-BO. This trade-off worsens with larger models.
By contrast, FT+L attempts to impose a constraint but falls short on S-ES, rendering it nearly unusable.

\textbf{MEND} displays divergent behavior by task.
For \textsc{ScEdit-CF}, S-ES drops by roughly 53\% compared to simpler PS tasks. It should be noted that WikiText-based training may contribute to the performance gap, yet the drastic drop remains noteworthy, especially since PS and S-ES share the same CounterFact edits. This suggests potential difficulties in adapting to different tasks.
In contrast, MEND remains comparatively more viable for \textsc{ScEdit-T}.

\textbf{ROME} achieves the best overall results across all models and all metrics, suggesting that a locate-then-edit strategy still offers strong performance in script-based scenarios. 

\textbf{MEMIT}, designed for large-scale editing, exhibits moderate S-ES but attains the highest S-NS and S-BO scores, indicating particularly strong preservation of unrelated facts.

Lastly, although \textbf{PROMPT} excels in S-ES for \textsc{ScEdit-T}, its less favorable locality metrics reveal limitations when handling script-based contexts.

\subsection{Results on Text-Level Metrics}

\subsubsection{Automatic Evaluation}
\label{Sec:automatic_evaluation}
We use GPT-4 to evaluate four text-level metrics on \textsc{ScEdit-CF} for LLAMA3-8B-generated scripts after editing.
While token-level metrics primarily capture edit performance, text-level metrics offer a more holistic assessment of how well a model integrates, generates, and reasons based on edited knowledge. Table~\ref{tab:text_metrics} and Figure~\ref{fig:Text-level} show the results. Additionally, we include text-level metrics for several large-scale closed-source commercial models in Appendix~\ref{sec:additional}.
\\[3pt]
\noindent\textbf{Coherence.} Coh. evaluates text-level edit effectiveness. PROMPT and ROME perform relatively well, aligning with their high S-ES scores. However, with 7 as the maximum score, these results remain unsatisfactory, highlighting that even token-level strong methods still have room for improvement at the text level. In contrast, FT nearly wipes out the model’s capabilities, fixating on the target object $o$ or even part of its tokens, which can hardly be considered an effective text-level edit.
% FT, on the other hand, shows the opposite trend. Direct fine-tuning nearly wipes out the model’s capabilities: despite extremely high token-level scores, it becomes fixated on producing the target object $o$ and even part of its tokens, which in many cases cannot be considered an effective edit at the text level.
\\[3pt]
\noindent\textbf{Executability and Completeness.} Exec. and Comp. do not directly assess the newly edited facts but rather probe whether the model’s inherent script-related capabilities remain intact following the edits.
MEMIT achieves the strongest performance here, possibly at the cost of Coh. 
ROME and PROMPT also perform well, with PROMPT even outperforming the Base Model in terms of Comp., suggesting that it remains largely unaffected in terms of interpreting the \textbf{Script Question} and providing a well-rounded response. 
By contrast, FT registers poor results again,  reflecting the irreparable damage it causes to the model’s broader script-related capabilities.
\\[3pt]
\noindent\textbf{Consistency.} Cons. checks whether the knowledge is stable, without mixing old and new facts. MEMIT and ROME both preserve consistency effectively, whereas PROMPT underperforms slightly here. This observation underscores that methods relying on in-context learning of the model can still face challenges in maintaining stable, conflict-free edits at the textual level.

\subsubsection{Human Evaluation}
Given the complexity of automated text-level evaluations, we further conduct a human evaluation on 400 sampled generated scripts. 
Three independent annotators, experienced in KE but uninvolved in the automated evaluation, scored the same four text-level metrics using the same criteria as GPT-4.
Krippendorff’s $\alpha$ of 0.43 and Spearman's $\beta$ of 0.72 (between human and automated measures) indicate moderate to substantial agreement. Detailed statistics and analysis are provided in Appendix~\ref{sec:human_evaluation}.

\subsection{Analysis of the Correlation of All Metrics}

Inspired by~\citet{rosati-etal-2024-long}, we analyze relationships between all metrics. GE\footnote{Weighted average of bi- and tri-gram entropies~\cite{zhang2018generating} employed by~\citet{meng2022locating} in the original ROME papers.} is included here. This represents a first attempt to integrate generative ability into KE evaluation. However, calculating entropy using short n-grams cannot fully capture the information present at the text level.

\begin{figure}
    \setlength{\belowcaptionskip}{-8pt} % 调整标题与下方内容之间的下边距
    \centering
    \includegraphics[width=1\linewidth]{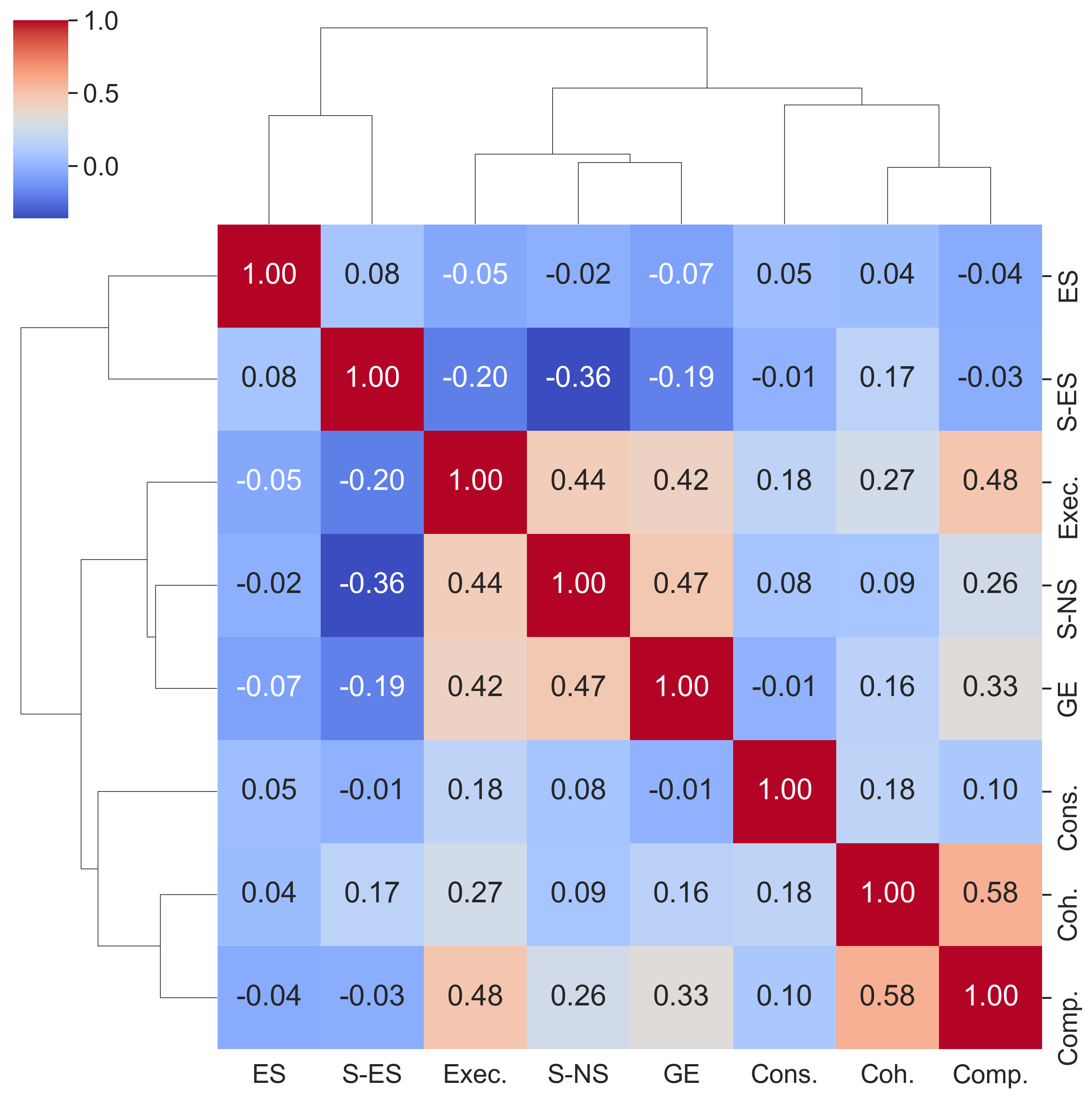}
    \caption{Clustered spearman correlation heatmap of token-level and text-level metrics}
    \label{fig:heatmap}
\end{figure}

Figure~\ref{fig:heatmap} presents a clustered Spearman correlation heatmap comparing token-level with text-level metrics. All statistically significant correlations (with $p < 0.05$ and $|\rho| > 0.1$) are detailed and further analyzed in Appendix~\ref{sec:analysis_correlation}.

In summary, our analysis yields three findings:
\begin{enumerate*}
    \item Fact-based efficacy (ES) alone fails to capture editing effectiveness in script scenarios.
    \item Combining Exec. and Comp. — which incorporate the script's inherent feature — provides a valuable complement to generative ability and specificity.
    \item Text-level edit effectiveness (Coh.) shows weak correlation with S-ES, while Cons. exhibits almost no relationship with token-level metrics, indicating that each captures distinct dimensions. We further illustrate this in a case study (Appendix~\ref{sec:case_study}), showcasing samples where token-level metrics are high but text-level edits face significant issues. Therefore, integrating metrics across levels may lead to a more comprehensive evaluation.
\end{enumerate*}

\vspace{-3mm}
\section{Discussion}

% \subsection{Analysis about the interrupted step}

Both text-level and token-level metrics reveal an inherent trade-off in \ds{} between achieving highly effective edits and limiting their broader impact on the model's performance. 

The deterministic nature of scripts enables a more definitive evaluation. Issues become more apparent at text level, highlighting the challenges of holistic editing in script-like scenarios, which require further research and advanced KE strategies.
% \subsubsection{Results when batch-editing}

\section{Conclusion}
In this paper, we present \ds{}, a novel script-based benchmark for evaluating KE methods in real-world scenarios. Through rigorous experiments, we highlight several limitations of current KE methods in handling script-based evaluations. Some methods like FT struggle to maintain Efficacy and Specificity. While methods like ROME achieve strong token-level performance, text-level scores reveal room for improvement in script-like scenarios. Further analysis between token-level and text-level metrics underscores the need for more comprehensive evaluation frameworks. We hope that \ds{} will inspire the development of more advanced KE techniques capable of addressing real-world complexities.
% \begin{figure}[!b]
%     \centering
%     \includegraphics[width=1\linewidth]{figs/Text-level_figure1.png}
%         \caption{Results of four text-level metrics. For clarity, the vertical axes for ``Exec.'' and ``Cons.'' begin at 6, while those for the other metrics start at 2.}
%     \label{fig:Text-level}
% \end{figure}
% \section{Limitation}

\section{Limitations}

\paragraph{Models.} We only edit a few basic LLMs, leaving many others unexplored. Additionally, due to resource limitations, the LLMs we edit have fewer than 10B parameters, excluding larger models. Moreover, several task-oriented planning LLMs remain untested.

\paragraph{Editing Methods.}In this paper, we primarily focus on comparing the effects of existing editing methods across different types of edits and evaluation granularities. However, the results leave room for improvement. Moving forward, our goal is to explore efficient and accurate editing across all granularities, especially at the text level. This may include investigating techniques like step-verifiers, which are commonly employed to improve language planning tasks~\cite{brahman2024plasma}, as well as other post-hoc methods. While we introduced \textbf{Script} scenarios, the editing methods themselves remain rooted in triple-level paradigms. Developing methods to support unstructured edits is a promising direction for future research. Furthermore, exploring scalability (massive and sequential editing capabilities) in \textbf{Script} scenarios represents another important avenue for advancement.

\paragraph{Automatic Evaluation.} Overestimation or underestimation may occur when doing automatic evaluation for generated texts~\cite{yuan-etal-2023-distilling}. To mitigate this, we incorporate moderate human evaluation and several correlation analyses.

\paragraph{Further Challenges} \ds{} is generated by GPT-4, potentially biasing it toward causal language models — a common issue with machine-generated data. Some incorrect or atypical samples emerge, though manual checks partially address this. What's more, expanding KE datasets to language planning domains may lead to some incompatibility or repetitive \textbf{Script Questions}, and the counterfactual edits may not incorporate well with real-world scenarios. Lastly, we focus on human-level script execution, leaving robot execution~\cite{lu2023neurosymbolic,huang2022language} unstudied, which highlights the challenges of translating complex human language into robot-executable forms and the gap toward embodied AI.

\section*{Acknowledgments}
This work is supported by the National Natural Science Foundation of China (Grant No. 62306087 and 62472121), the Natural Science Foundation of Shandong Province (Grant No. ZR2023QF154), Special Funding Program of Shandong Taishan Scholars Project and CCF-Sangfor `Yuanwang' Research Fund (Grant No. 20240201).

\bibliography{custom}

\appendix
\begin{figure*}[]
    \centering
    \includegraphics[width=\linewidth]{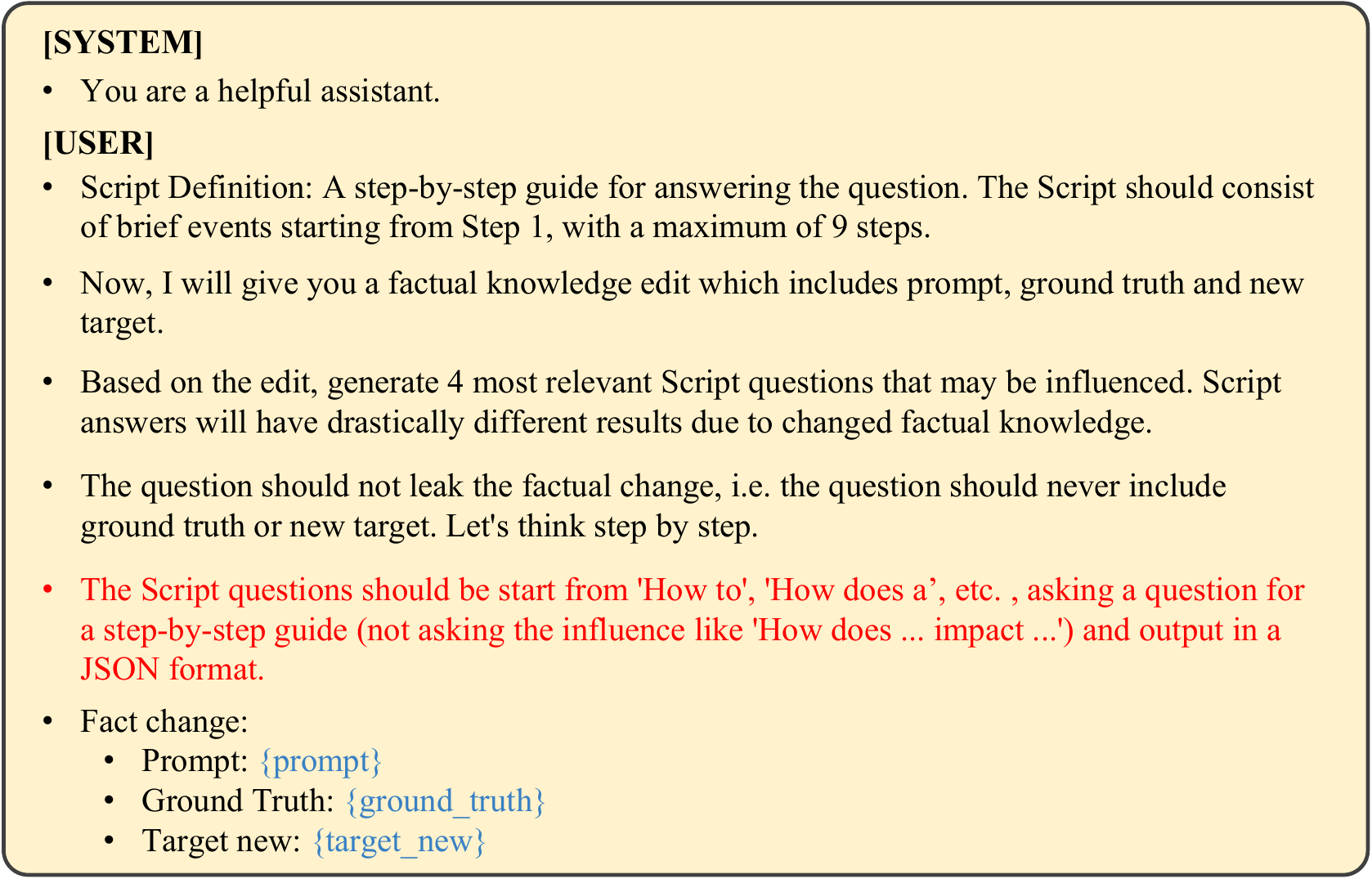}
    \caption{Prompts for generating \textbf{Script Questions} that are significantly influenced by the knowledge updates.}
    \label{fig:script_question_generation}
\end{figure*}
\begin{figure*}
    \centering
    \includegraphics[width=\linewidth]{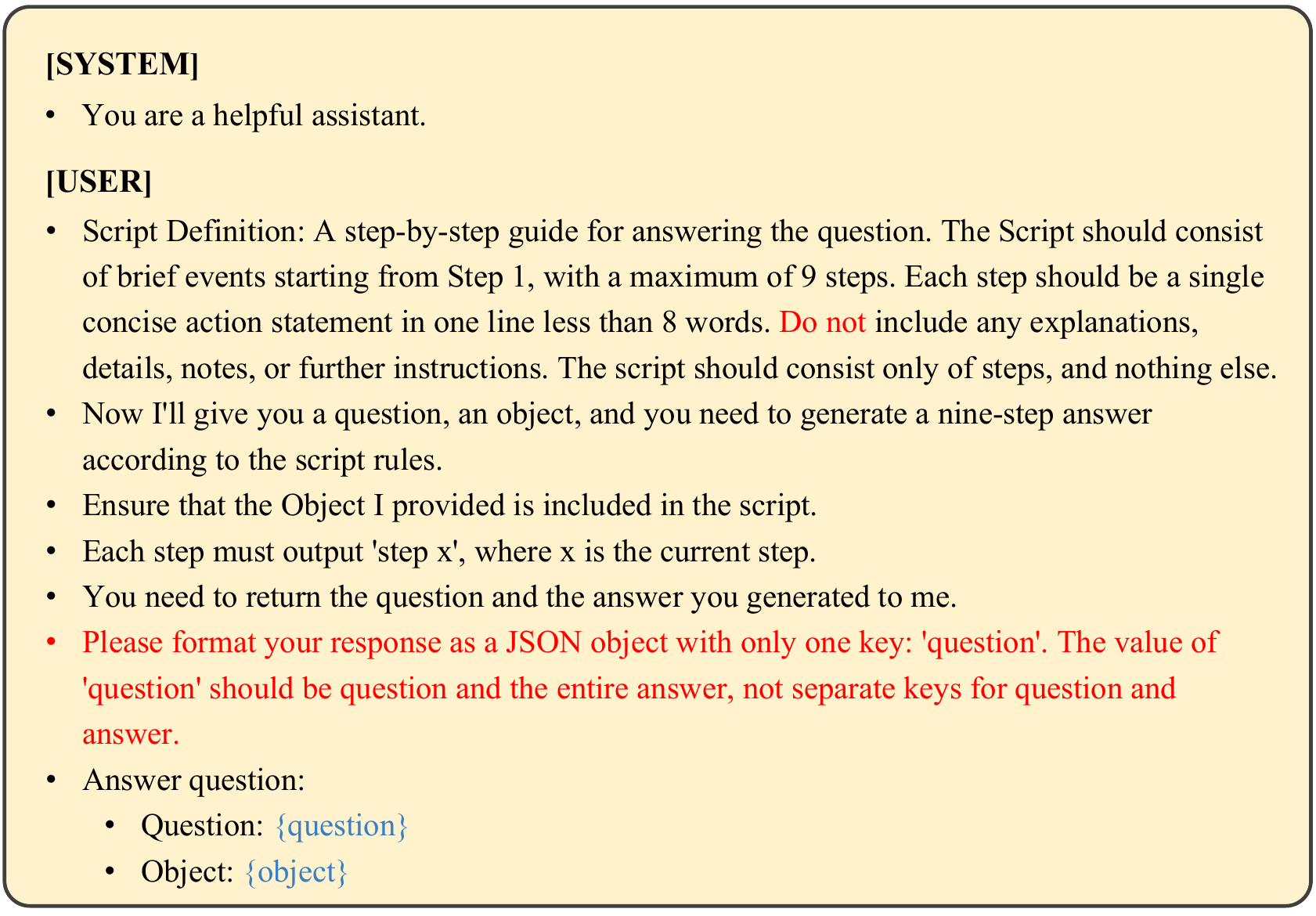}
    \caption{Prompts for generating \textbf{Scripts} based on given \textbf{Script Questions}.}
    \label{fig:script_generation}
\end{figure*}

\section{Scripts Format Details}
\label{sec:Script_format}
In \ds{}, we define the \textbf{Script} in a specific format as follows:
\begin{itemize*}
    \item A step-by-step guide for answering the \textbf{Script Question}.
    \item Consists of brief events starting from Step 1, with a maximum of 9 steps.
    \item Each step should be a single concise action statement in one line less than 8 words.
    \item Excludes any explanations, details, notes, or additional instructions.
    \item The script should contain only the steps, and nothing else.
\end{itemize*}

We prompt the model with following prompt when executing text-level evaluation:
\begin{lstlisting}
Provide a step-by-step guide in Script form for answering the question: {Script Question}. 
The Script should consist of brief events starting from Step 1, with a maximum of 9 steps. 
Each step should be a single concise action statement in one line less than 8 words. 
Do not include any explanations, details, notes, or further instructions. 
The script should consist only of the steps, and nothing else.

\end{lstlisting}

\section{Dataset Construction Details}
\label{sec:Dataset_Construction}
\subsection{Data Construction}
We leverage \textit{gpt-4o-2024-11-20} to construct \textsc{ScEdit-CF} as well as \textsc{ScEdit-T}.

Two main processes are utilized in our dataset construction. First, as illustrated in Figure~\ref{fig:script_question_generation}, we prompt GPT-4 with few-shot examples to generate \textbf{Script Questions} that are significantly influenced by the edits. Second, shown in Figure~\ref{fig:script_generation}, we prompt GPT-4 with specific requirements to generate \textbf{Scripts} to response these \textbf{Script Questions} based on old knowledge. These \textbf{Scripts} are then truncated and concatenated with the corresponding \textbf{Script Questions} to form a cloze-format, script-based prompts aligning with token-level evaluation paradigms. Neighboring facts are also collected and processed similarly.

\subsection{Data Filtering and Quality Evaluation}
\label{ssec:Data_Filtering}

During the dataset creation process, we conducted comprehensive filtering and verification to ensure data quality. The primary focus of our filtering process was to remove data points that were either difficult to edit or evaluate or where data leakage was identified.

\subsubsection{Types of Data Filtered Out}
We addressed several key issues:

\begin{itemize}
    \item \textbf{Hallucination in GPT-4-generated Data:} In some instances, GPT-4-generated data exhibited hallucinations where the original object was missing from the script, rendering truncation impossible. These cases were identified and removed before further evaluation.

    \item \textbf{Leakage of Original Object-related Knowledge in Truncated Scripts:} This issue was particularly critical in \textsc{ScEdit-CF}, where original knowledge is transformed into counterfactual knowledge. Even when truncated before the original object, scripts occasionally retained related knowledge, which could confuse the model. For example, in the edit \texttt{(Panamera, manufactured by, Porsche $\rightarrow$ Ford)}, if the truncated script still included knowledge such as “headquartered in Stuttgart, Germany,” it could create ambiguity. To address this, we utilized GPT-4 as a verifier to identify ambiguous cases, which were then manually filtered out.

    \item \textbf{Truncated Prompts Where the Target Object Appears Too Early:} In some cases, the target object appeared too early in the truncated prompt, making it challenging to generate the desired target object. Statistical analysis revealed that 6.75\% of the data was truncated at Step 1. These cases underwent manual filtering.
\end{itemize}
\begin{table}[h!]
    \centering
    \resizebox{\columnwidth}{!}{
    \begin{tabular}{|p{2.5cm}|p{6.5cm}|}
        \hline
        \textbf{Fact} & (Eiffel Tower, location, Paris) \\ \hline
        \textbf{Edit} & Paris $\rightarrow$ Rome \\ \hline
        \textbf{Fact Prompt} & The Eiffel Tower is in () \\ \hline
        \textbf{Script Question $Q_{i,k}$} & How can a Chinese travel to the Eiffel Tower? \\ \hline
        \textbf{Generated Script} &
        Step 1: Open Trip.com. \newline
        Step 2: Search for international flights. \newline
        Step 3: Select your preferred flight option. \newline
        Step 4: Buy a flight ticket to \big| Paris Airport. \newline % Note: \big| might need review for its purpose here. Usually for delimiters.
        Step 5: Apply for a Schengen visa. \newline
        Step 6: Book a hotel in Paris. \newline
        Step 7: Fly to France and reach your hotel. \newline
        Step 8: Visit the Eiffel Tower. \newline
        Step 9: Take a photo with the Eiffel Tower. \\ \hline
        \textbf{Script-based Prompt $\widetilde{Q_{i,k}}$} & How can a Chinese travel to the Eiffel Tower? \newline
        Step 1: Open Trip.com. \newline
        Step 2: Search for international flights. \newline
        Step 3: Select your preferred flight option. \newline
        Step 4: Buy a flight ticket to () \\ \hline
        \textbf{Neighbor Fact} & (Louvre Museum, location, Paris) \\ \hline
        \textbf{Script-based Neighborhood Prompt $\widetilde{Q_{i,k}}^\prime$} & How does a tourist in Korea visit the Louvre Museum? \newline
        Step 1: Apply for a Schengen visa (if required). \newline
        Step 2: Book tickets for the Louvre Museum. \newline
        Step 3: Select your preferred date and time. \newline
        Step 4: Book a flight ticket to () \\ \hline
    \end{tabular}
    }
    \caption{An example of the constructed data (showcasing only one question and one neighbor fact here).}
    \label{tab:data_example}
\end{table}
\subsubsection{Filtering and Verification Process}
Our filtering process involved multiple stages, combining automated checks, GPT-4-assisted checks, and manual verification. Overall, 8.23\% of the GPT-4-generated data was filtered out during this process.

To further ensure data quality, we randomly sampled 10\% of the final dataset for manual evaluation by three experts. The experts evaluated the data based on the following criteria:
\begin{itemize}
    \item Whether the generated Script Questions were meaningful.
    \item Whether the model's responses to the Script Questions showed significant changes before and after editing.
    \item Whether the truncation positions effectively guided the generation of the target object.
\end{itemize}
The evaluation achieved an average agreement rate of 94.5\%, which supports the quality of the dataset.

\subsection{Data Example}
We present an example of the constructed data in Table~\ref{tab:data_example}, which includes a specific script question $Q_{i,k}$, a cloze-format script-based prompt $\widetilde{Q_{i,k}}$, and a cloze-format script-based neighbor prompt $\widetilde{Q_{i,k}}^\prime$.

\begin{figure*}
    \centering
    \includegraphics[width=\linewidth]{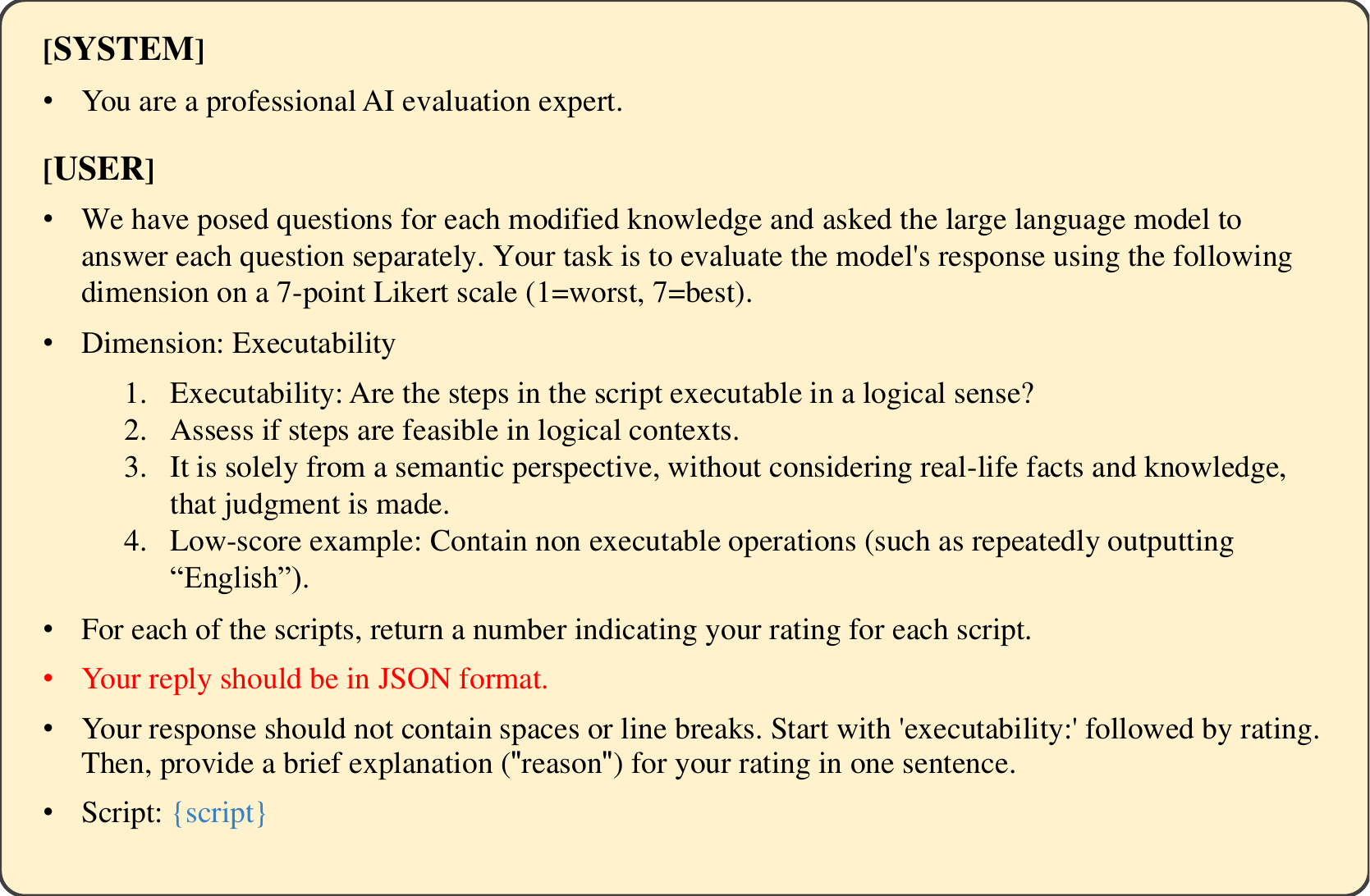}
    \caption{Prompts for evaluating Executability.}
    \label{fig:executability}
\end{figure*}
\begin{figure*}
    \centering
    \includegraphics[width=\linewidth]{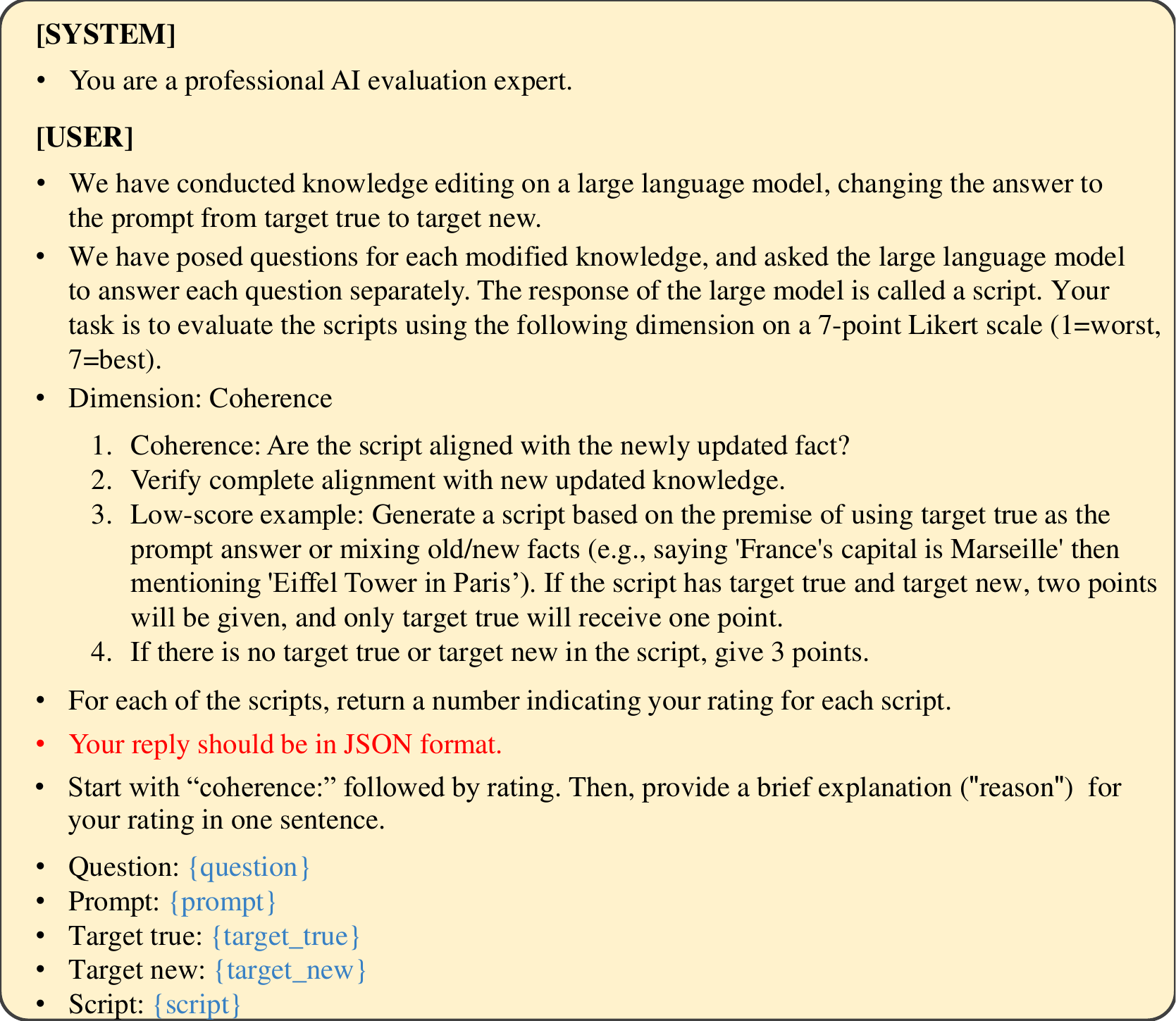}
    \caption{Prompts for evaluating Coherence.}
    \label{fig:coherence}
\end{figure*}
\begin{figure*}
    \centering
    \includegraphics[width=\linewidth]{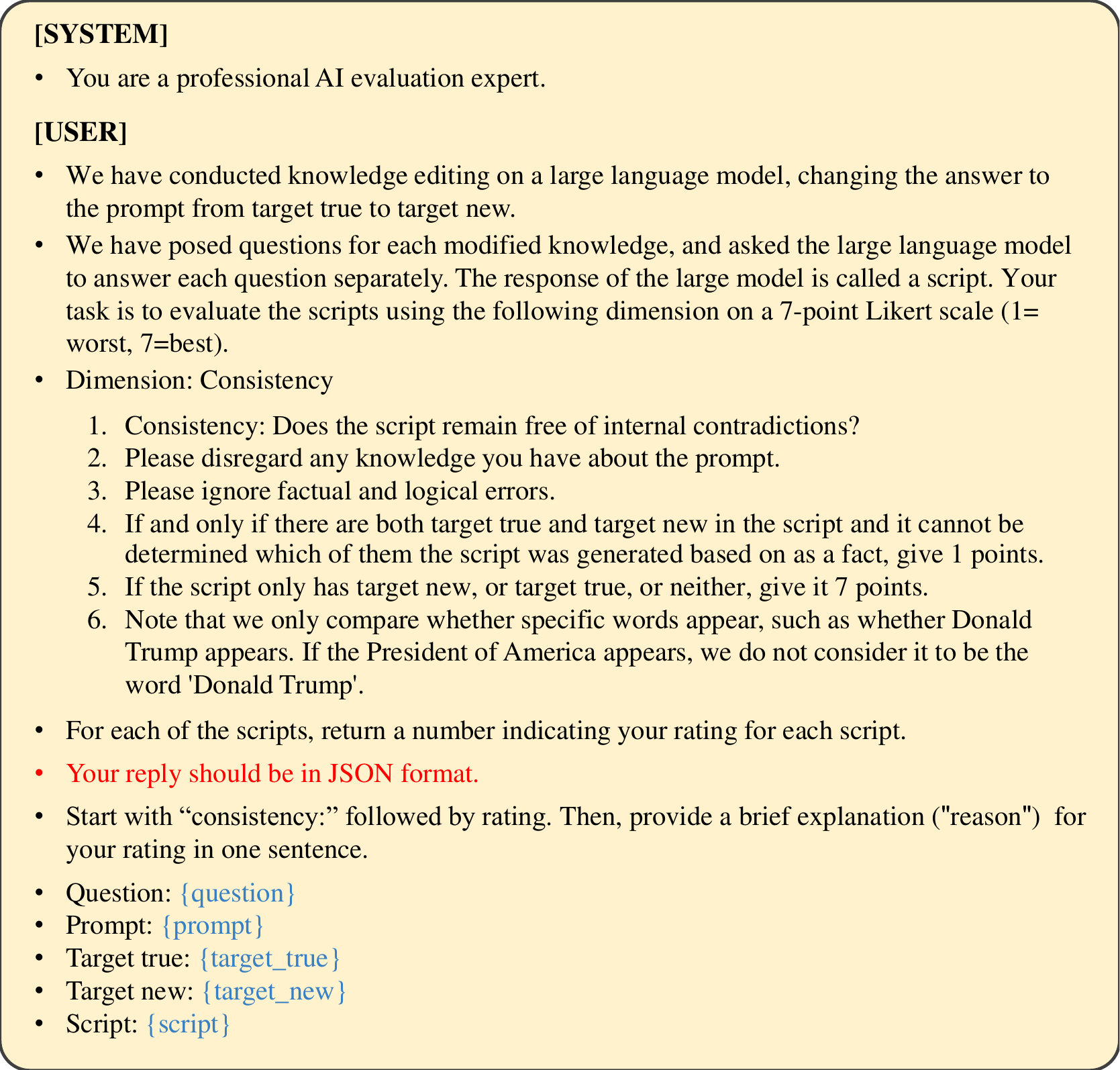}
    \caption{Prompts for evaluating Consistency.}
    \label{fig:consistency}
\end{figure*}

\begin{figure*}
    \centering
    \includegraphics[width=\linewidth]{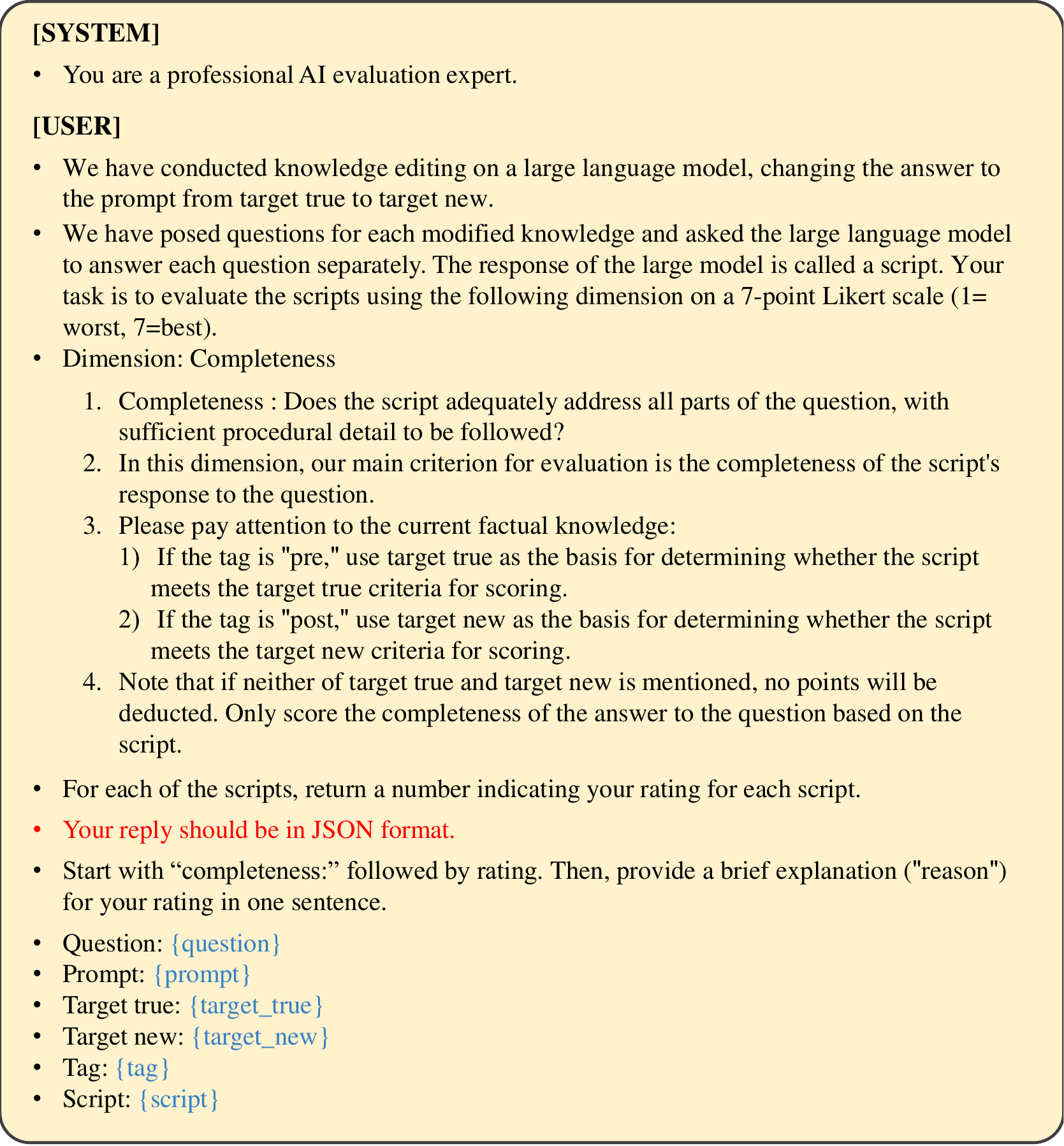}
    \caption{Prompts for evaluating Completeness.}
    \label{fig:completeness}
\end{figure*}
\section{Text-level Evaluation Details}
% \label{sec:text-level details}

\subsection{Evaluation Criterion and Prompts}
\label{sec:text-level details}

We employ \textit{gpt-4o-2024-11-20} with a few-shot approach to automatically evaluate text-level metrics in \textsc{ScEdit-CF}.

Figures~\labelcref{fig:executability,fig:coherence,fig:consistency,fig:completeness} present the evaluation criteria and carefully crafted prompts used in the evaluation process. For the sake of clarity, several few-shot examples were omitted.

\newpage

\subsection{Human Evaluation}
\label{sec:human_evaluation}
\begin{table}[]
    
    \setlength{\belowcaptionskip}{-17pt} % 调整标题与下方内容之间的下边距
    \centering
    \begin{tabular}{lc}
    \toprule
    \textbf{Metric} & \textbf{Krippendorff’s $\alpha$} \\
    \midrule
    Executability & 0.59 \\
    Coherence & 0.35\\
    Consistency & -0.09\\
    Completeness & 0.25 \\
    \bottomrule
    \end{tabular}
    \caption{Metric-wise results of inter-rater agreement between annotators.}
    \label{tab:metric_wise}
\end{table}
We conducted a human evaluation with the help of three researchers experienced in KE, who were not involved in the automated evaluation process. The inter-rater Krippendorff’s Alpha coefficient indicates moderate agreement ($\alpha = 0.45$). Detailed metric-wise results are presented in Table~\ref{tab:metric_wise}, where some metrics, such as Exec. and Coh., show higher agreement, while Cons. exhibits poor agreement, reflecting its subjective nature. 

The human evaluation results align closely with the findings and conclusions of (\S \ref{Sec:automatic_evaluation}). We acknowledge~\citet{rosati-etal-2024-long} and believe one metric with poor agreement does not undermine our overall findings, especially considering the use of a seven-point Likert scale. However, it still highlights the need for further exploration of editing stability in future research. 

For comparisons between human and automatic measures, Krippendorff’s $\alpha$ of 0.43 and Spearman’s $\beta$ of 0.72 indicate moderate to substantial agreement, suggesting that automated scores align reasonably well with human raters. Increasing the number of examples under few-shots settings could be a direction for future improvements.

\section{Experiment Setup}

Our experiments build upon ROME~\cite{meng2022locating} and MEMIT~\cite{meng2022memit}. We additionally incorporate the LLAMA3 covariance matrix provided by~\citet{fang2024alphaeditnullspaceconstrainedknowledge}, and adapt implementation details of the PROMPT method from~\citet{ammar-khodja-etal-2024-wikifactdiff-large}.

All experiments are conducted on a single A40 (48GB) GPU. The LLMs are loaded using HuggingFace Transformers~\cite{wolf-etal-2020-transformers}. 

In \ds{}, each edit is performed individually, focusing on the specified $(s,r,o)$ triple in a single operation rather than in a sequential chain. After each evaluation, the edited model is reverted to its original state, ensuring that edits remain isolated and do not affect subsequent operations. This design allows for a controlled script-based assessment of each individual KE.

\section{Detailed Analysis of the Correlation of All Metrics}
\label{sec:analysis_correlation}
\begin{table}[ht]
\centering
\begin{tabular}{l c}
\toprule
\textbf{Metric Pair} & \textbf{$\rho$} \\
\midrule
\multicolumn{2}{l}{\textbf{Text-level Metrics}} \\
Coh. vs. Comp.      & 0.58 \\
Exec. vs. Comp.     & 0.48 \\
Exec. vs. Coh.      & 0.27 \\
Exec. vs. Cons.     & 0.18 \\
Coh. vs. Cons.      & 0.18 \\
\midrule
\multicolumn{2}{l}{\textbf{Token-level Metrics}} \\
S-NS vs. GE           & 0.47 \\
S-NS vs. S-ES         & -0.36 \\
S-ES vs. GE         & -0.19 \\
\midrule
\multicolumn{2}{l}{\textbf{Token-level vs. Text-level Metrics}} \\
S-NS vs. Exec.        & 0.44 \\
GE vs. Exec.        & 0.42 \\
GE vs. Comp.        & 0.33 \\
S-NS vs. Comp.        & 0.26 \\
S-ES vs. Exec.      & -0.20 \\
S-ES vs. Coh.       & 0.17 \\
GE vs. Coh.         & 0.16 \\
\bottomrule
\end{tabular}
\caption{Statistically significant ($p < 0.05$) combined Spearman's rank correlations for metric pairs with $|\rho|>0.1$.}
\label{tab:combined_correlations}
\end{table}

In this section, we present a thorough analysis of the correlations among various performance metrics computed at both the text and token levels.
Table~\ref{tab:combined_correlations} summarizes all the statistically significant correlations (with $p < 0.05$ and $|\rho|>0.1$) observed in our analysis.

\subsection{Text-Level Metrics}
Among the text-level metrics, the Comp. exhibits moderate to strong correlations with both Exec. and Coh., with Spearman's rank correlation coefficients of $\rho = 0.48$ and $\rho = 0.58$, respectively. Exec. and Coh. shows a weak positive correlation ($\rho = 0.27$), suggesting a weak association between the editing effectiveness and the inherent script-based generation ability. This relationship, which may appear counterintuitive, could be attributed to the fine-tuning (FT) effects discussed in (\S \ref{Sec:automatic_evaluation}). Furthermore, Cons. shows weak or negligible correlations with all other text-level metrics, implying that it likely captures a unique aspect of performance that is not reflected in the other measures.

\subsection{Token-Level Metrics}
At the token level, ES does not show a significant correlation with either the S-ES or S-NS. However, a moderate negative correlation exists between S-NS and S-ES ($\rho = -0.36$). This negative relationship indicates that relying solely on Fact Edit Efficacy may be insufficient in script scenarios. Additionally, GE exhibits a moderate positive correlation with S-NS ($\rho = 0.47$), suggesting an intuitive link between generative ability and the specificity

\subsection{Cross-Level Correlations}
Beyond the intra-level correlations, cross-level analysis reveals several interesting patterns. Notably, Exec. correlates moderately with both GE ($\rho = 0.42$) and S-NS ($\rho = 0.44$). Moreover, GE shows a moderate correlation with Comp. ($\rho = 0.33$). In contrast, S-ES is only weakly correlated with the Coh. ($\rho = 0.17$), and S-NS shows a weak correlation with Comp. ($\rho = 0.26$). This time, S-ES exhibits a weak but intuitive negative correlation with Exec. ($\rho = -0.20$).

\begin{table*}[!t]
\centering
\resizebox{2\columnwidth}{!}{
\begin{tabular}{|l|l|}
\hline
\textbf{Fact} & \textbf{(Thomas Joannes Stieltjes, mother tongue, \textcolor{dataset_red}{Dutch})} \\
\hline
\textbf{Fact Prompt} & \textbf{The mother tongue of Thomas Joannes Stieltjes is}   \\ \hline
\textbf{Target True} & \textbf{\textcolor{dataset_red}{Dutch}}  \\ \hline
\textbf{Target New} & \textbf{\textcolor{dataset_green}{English}}   \\ \hline
\textbf{Script Question} & \textbf{How to discover the works of Thomas Joannes Stieltjes in his mother tongue?}   \\ 
\hline
\textbf{Base Script} & \textbf{Evaluation} \\ 
Step 1: Open Google search engine.  & (Exec: 7, Coh: 1, Cons: 7, Comp: 7)\\
Step 2: Type "Stieltjes werken" in search bar. & \hspace{0.5cm} \multirow{8}{*}{\includegraphics[width=4.2cm]{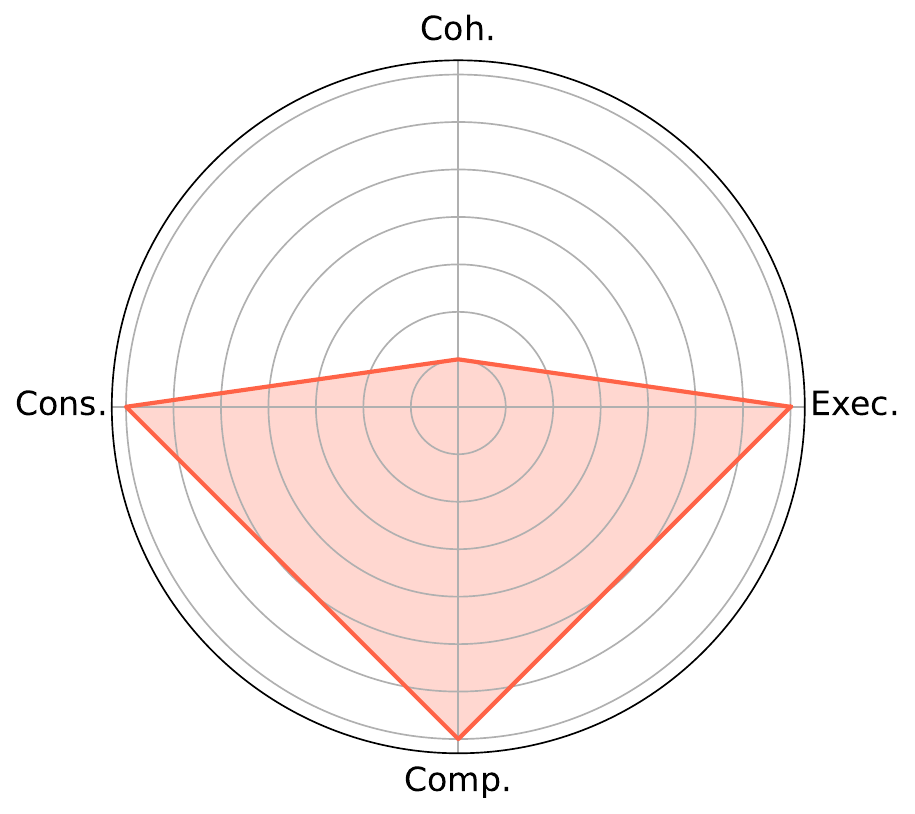} } \\
Step 3: Press Enter key.  & \\
Step 4: Browse search results for "Stieltjes werken".  & \\
Step 5: Click on "{\textcolor{dataset_red}{\underline{Nederlandse Taal}}}" link. & \\
Step 6: Scroll down to "Werken" section. & \\
Step 7: Click on "PDF" link for "Recherches sur & \\ 
les fractions continues". & \\
Step 8: Download PDF file. & \\
Step 9: Open PDF file in Adobe Acrobat.  & \\ \hline
\textbf{FT Script} & \textbf{Evaluation} \\ Step 1: Search Google for "\textcolor{dataset_green}{English} \textcolor{dataset_green}{English} \textcolor{dataset_green}{English}". & (Exec: 1, Coh: 3,  Cons: 7,  Comp: 1)\\
Step 2: Click on the search results link. & \hspace{0.5cm} \multirow{7}{*}{\includegraphics[width=4.2cm]{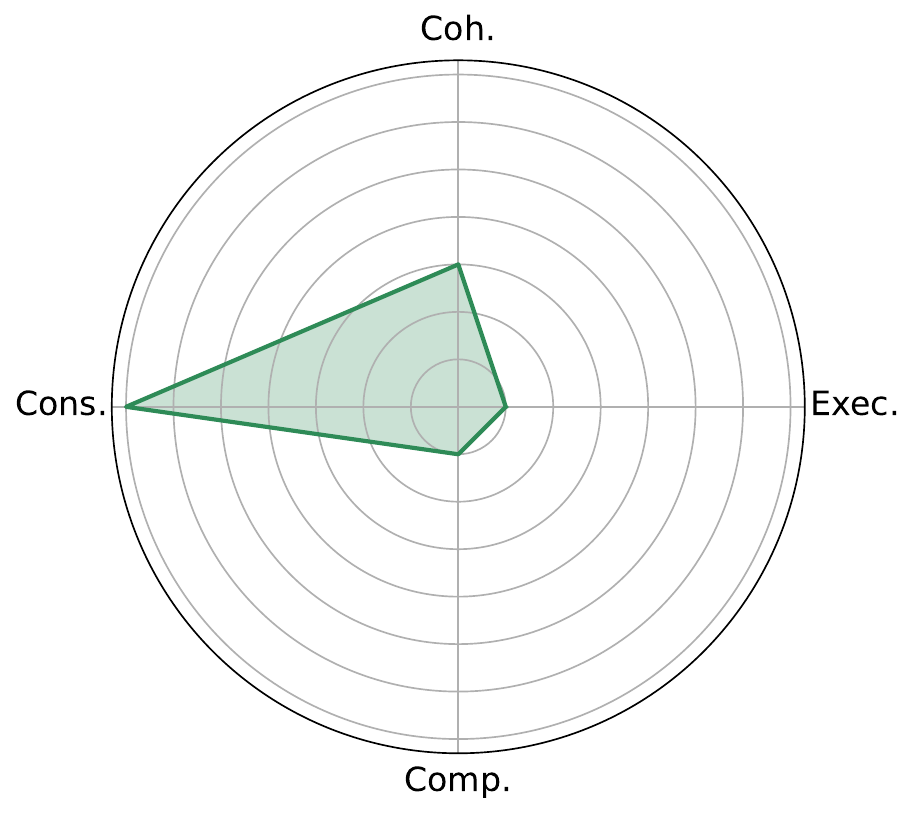}}\\
Step 3: Open the website "\textcolor{dataset_green}{English} \textcolor{dataset_green}{English} \textcolor{dataset_green}{English}". &  \\  
Step 4: Click on the "\textcolor{dataset_green}{English} \textcolor{dataset_green}{English} \textcolor{dataset_green}{English}" tab. & \\
Step 5: Click on the "\textcolor{dataset_green}{English} \textcolor{dataset_green}{English} \textcolor{dataset_green}{English}" link. & \\
Step 6: Click on the "\textcolor{dataset_green}{English} \textcolor{dataset_green}{English} \textcolor{dataset_green}{English}" link. & \\
Step 7: Click on the "\textcolor{dataset_green}{English} \textcolor{dataset_green}{English} \textcolor{dataset_green}{English}" link. & \\
Step 8: Click on the "\textcolor{dataset_green}{English} \textcolor{dataset_green}{English} \textcolor{dataset_green}{English}" link. & \\
Step 9: (repeated entries for \textcolor{dataset_green}{English}...) & \\ \hline
\textbf{ROME Script} & \textbf{Evaluation} \\ Step 1: Open Google search engine. & (Exec: 6,  Coh: 2,  Cons: 7,  Comp: 2)\\
Step 2: Type "Thomas Joannes Stieltjes in \textcolor{dataset_blue}{Arabic}". & \hspace{0.5cm} \multirow{7}{*}{\includegraphics[width=4.2cm]{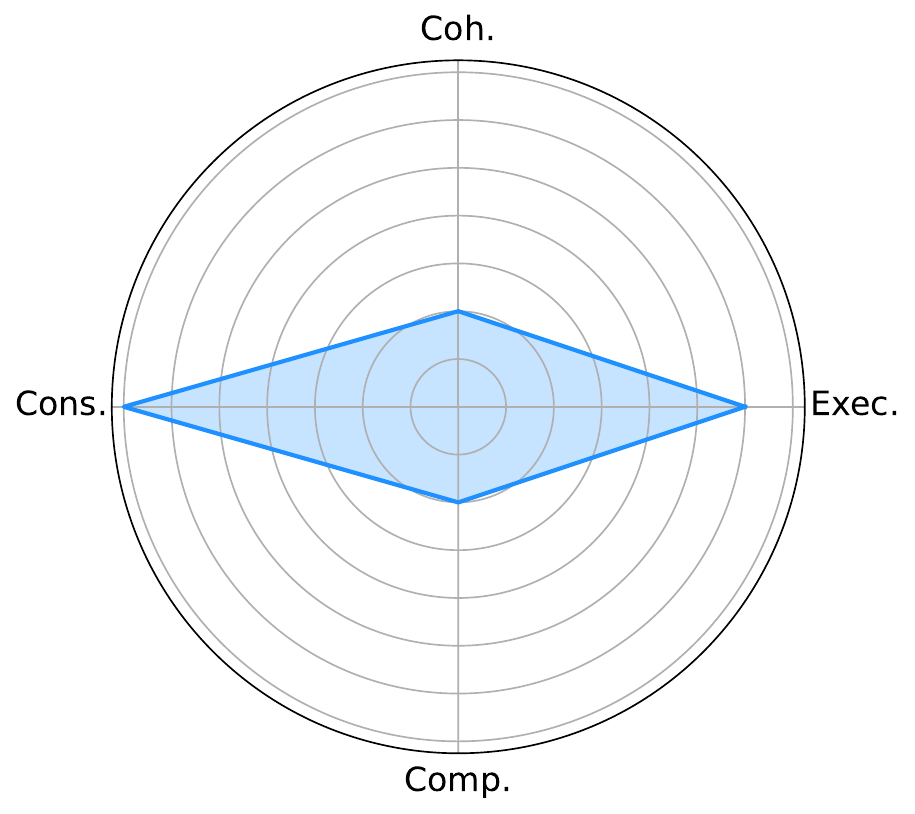}} \\
Step 3: Press Enter key. & \\
Step 4: Review search results. & \\
Step 5: Look for official website. & \\
Step 6: Check for \textcolor{dataset_blue}{Arabic} language option. & \\
Step 7: Select \textcolor{dataset_blue}{Arabic} language option. & \\
Step 8: Read his works in \textcolor{dataset_blue}{Arabic}. & \\
Step 9: Bookmark his website. & \\ \hline
\textbf{MEMIT Script} & \textbf{Evaluation}\\ Step 1: Search online for Stieltjes' biography. & (Exec: 7,  Coh: 7,  Cons: 7, Comp: 7)\\
Step 2: Check the biography for native language. & \hspace{0.5cm} \multirow{7}{*}{\includegraphics[width=4.2cm]{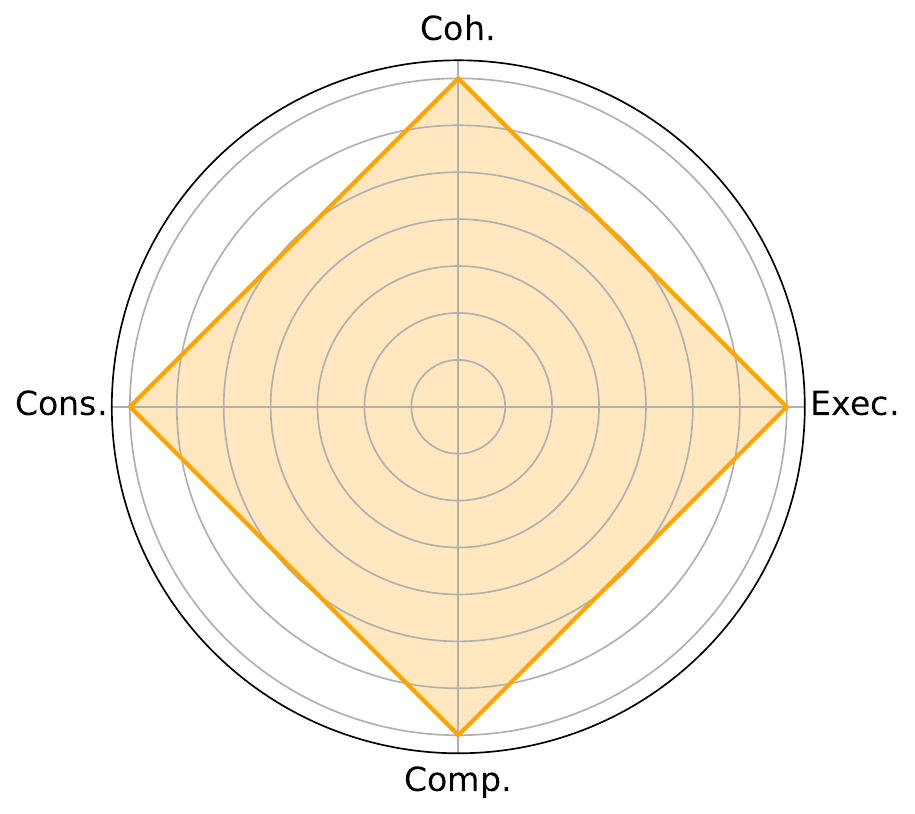}} \\
Step 3: Identify the native language as \textcolor{dataset_green}{English}. & \\
Step 4: Search online for Stieltjes' works in \textcolor{dataset_green}{English}. & \\
Step 5: Visit online archives of scientific journals. & \\
Step 6: Search journals for Stieltjes' publications. & \\
Step 7: Filter results by date of publication. & \\
Step 8: Download relevant articles in PDF format. & \\
Step 9: Read and analyze the articles. & \\ \hline
\textbf{PROMPT Script} & \textbf{Evaluation}\\ Step 1: Find Thomas Joannes Stieltjes' birthplace. & (Exec: 7,  Coh: 2,  Cons: 1,  Comp: 4)\\
Step 2: Research \textcolor{dataset_red}{Dutch} archives in that location. & \hspace{0.5cm} \multirow{7}{*}{\includegraphics[width=4.2cm]{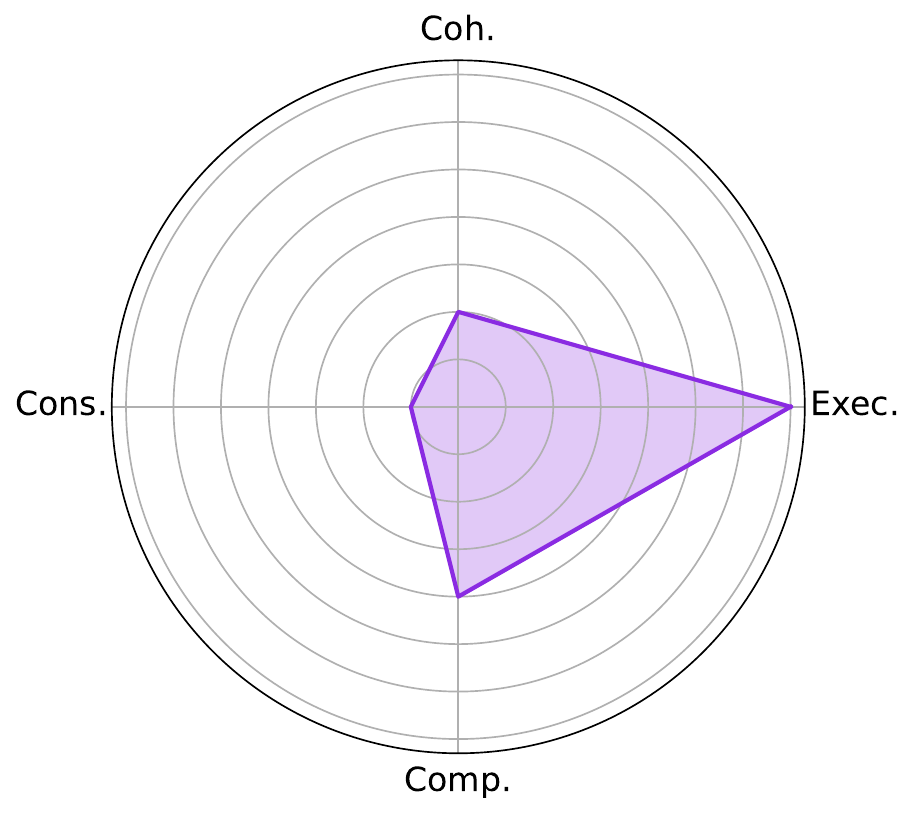}}\\
Step 3: Search for Stieltjes' writings in \textcolor{dataset_red}{Dutch}. & \\
Step 4: Visit libraries and museums in the Netherlands. & \\
Step 5: Consult experts in \textcolor{dataset_red}{Dutch} literature and history. & \\
Step 6: Check online archives and databases. & \\
Step 7: Translate relevant documents into \textcolor{dataset_green}{English}. & \\
Step 8: Read and analyze Stieltjes' works in \textcolor{dataset_green}{English}. & \\
Step 9: Share findings with the academic community. & \\ \hline
\end{tabular}

}
\caption{Case Study of our proposed text-level metrics. In the table, \textcolor{dataset_red}{red} indicates the original facts, \textcolor{dataset_green}{green} denotes the edited facts, and \textcolor{dataset_blue}{blue} represents all other unrelated facts.}
\label{table:stieltjes_script}
\end{table*}
\section{Case Study}
\label{sec:case_study}
We present a case study of the generated Script as shown in Table~\ref{table:stieltjes_script}. In the table, red indicates the original facts, green denotes the edited facts, and blue represents all other facts. 

In the \textbf{Base} Script, the term ``Nederlandse Taal'' is mentioned—this is originally in Dutch and translates into English as “Dutch Language.” Consequently, we rated its Coh. as 1 (Strongly Disagree), while awarding full marks for other metrics.

\textbf{FT} Script, however, simply repeats “English” in a loop. Although this complies with the editing directive, we do not consider it text-level coherent with the edited knowledge, so its Coh. received a score of 3 (Slightly Disagree). Moreover, it achieved the poorest performance in both Exec. and Comp.

\textbf{ROME}, which has shown great performance in our experiments, exhibits an unusual failure case here: rather than reflecting either the original or the updated knowledge, it introduces “Arabic.” This suggests that the editing direction may have gone awry. Notably, when we reviewed the token-level metrics, all metrics performed well. This case clearly underscores the importance of integrating text-level evaluation metrics. In light of this, we assigned a score of 2 (Disagree) for both Coh. and Comp.

\textbf{MEMIT} performed well in this instance—not only did it successfully incorporate the new knowledge into the generated Script, but it also maintained the performance of other Script-related aspects, earning full marks.

Finally, \textbf{PROMPT} demonstrated an instance of unstable editing. Instead of recognizing that Thomas Joannes Stieltjes’ mother tongue has been updated to English, the model interpreted this as a directive to translate into English. As a result, it received 1 (Strongly Disagree) point for Cons. and 2 (Disagree) points for Coh., with Comp. scoring 4 (Neutral/Uncertain).

\section{Additional Results of Commercial Closed-source Models}
\label{sec:additional}
Due to the limited availability of computational resources and restricted access to the weights of proprietary large-scale language models, most existing studies have predominantly focused on open-source foundational models. In alignment with prior work, our primary evaluations have covered the majority of open-source LLMs commonly employed in the literature. Nevertheless, we acknowledge the importance of broadening the evaluation scope to encompass a more diverse set of models, including closed-source, large-scale, and commercially mainstream systems.

To this end, we further extend our benchmark by evaluating knowledge editing (KE) methods on several representative closed-source models, including \textsc{GPT-4o}, \textsc{DeepSeek-v3}, and \textsc{Claude 3.5}. Specifically, we employ three methods: (1) PROMPT, following the same configuration as described in our main experiments; (2) IKE (In-context Knowledge Editing)~\cite{IKE}; and (3) Script-based IKE, which incorporates Script scenarios into the context to facilitate editing.

These methods were tested on a subset of the ScEdit-CF benchmark (100 cases), and performance was assessed using standard text-level evaluation metrics. The experimental results are summarized as follows (Table \ref{tab:closed_models}):

\begin{table}[ht]
\centering

\resizebox{\linewidth}{!}{
\begin{tabular}{lcccc}
\toprule
\textbf{Model} & \textbf{Exec.$\uparrow$} & \textbf{Coh.$\uparrow$} & \textbf{Cons.$\uparrow$} & \textbf{Comp.$\uparrow$} \\
\midrule
\multicolumn{5}{l}{\textbf{GPT-4o}} \\
\quad Base Model         & 7.00 & 2.82 & 7.00 & 6.34 \\
\quad PROMPT             & 6.96 & 4.55 & 6.95 & 5.82 \\
\quad IKE                & 6.88 & 5.06 & 6.96 & 6.18 \\
\quad Script-based IKE   & 6.86 & 5.44 & 6.92 & 6.36 \\
\midrule
\multicolumn{5}{l}{\textbf{DeepSeek-V3}} \\
\quad Base Model         & 7.00 & 2.96 & 7.00 & 6.33 \\
\quad PROMPT             & 6.88 & 4.84 & 6.90 & 5.88 \\
\quad IKE                & 6.76 & 5.13 & 6.85 & 6.18 \\
\quad Script-based IKE   & 6.82 & 5.69 & 6.88 & 6.36 \\
\midrule
\multicolumn{5}{l}{\textbf{Claude-3.5}} \\
\quad Base Model         & 7.00 & 2.22 & 6.99 & 6.16 \\
\quad PROMPT             & 6.58 & 3.65 & 6.56 & 4.63 \\
\quad IKE                & 5.56 & 3.32 & 6.57 & 3.27 \\
\quad Script-based IKE   & 6.16 & 6.06 & 6.94 & 5.49 \\
\bottomrule
\end{tabular}
}
\caption{Performance of KE Methods on Closed-Source Commercial LLMs.}
\label{tab:closed_models}
\end{table}
We summarize our key observations as follows:

\begin{itemize}
    \item \textbf{Effectiveness of ICL-based editing.} In-context learning (ICL) based editing methods remain effective on certain closed-source LLMs. On \textsc{GPT-4o} and \textsc{DeepSeek-V3}, all three editing strategies---\textsc{Prompt}, \textsc{IKE}, and \textsc{Script-based IKE}---consistently led to significant performance improvements across multiple metrics.

    \item \textbf{Contextual richness matters.} The form and richness of the contextual input play a crucial role in the effectiveness of editing. Moving from \textsc{Prompt} to \textsc{IKE} and then to \textsc{Script-based IKE}, the contextual information becomes progressively more structured and informative. This often leads to more effective script-based edits, demonstrating the ability of large models to learn and apply knowledge modifications through well-designed context.

    \item \textbf{Resistance in certain models.} Some LLMs, such as \textsc{Claude-3.5}, exhibit noticeable resistance to counterfactual knowledge editing. This may be attributed to stronger safety alignment mechanisms that actively reject edits perceived as misinformation. In particular, the editing performance of both \textsc{Prompt} and \textsc{IKE} on \textsc{Claude-3.5} was markedly lower than on other models. Nevertheless, after incorporating \textsc{Script-based IKE}, performance improved significantly, making it the best-performing method (\textsc{SOTA}) under our evaluation settings. This indicates both the model’s learning capacity and its unique alignment characteristics.
\end{itemize}

\noindent\textbf{Note.} The numerical results in Table~\ref{tab:closed_models} should not be directly compared with those in Table~\ref{tab:text_metrics}, as commercial models such as \textsc{GPT-4o}, \textsc{DeepSeek-V3}, and \textsc{Claude} possess much larger model scales and significantly stronger inherent reasoning capabilities. Comparing them directly with 7B-scale open-source models is not meaningful due to fundamental differences in architecture and capacity.
\end{document}